\documentclass[journal]{IEEEtran}
\usepackage{amsmath}
\usepackage{amssymb}

\usepackage{bbding}
\usepackage{amsmath}
\usepackage{amssymb}
\usepackage{array}
\usepackage{booktabs}
\usepackage{multirow}
\usepackage{multirow}
\usepackage{setspace}
\usepackage{color,subfigure}
\usepackage{mathrsfs}
\usepackage{graphics}
\usepackage{epsfig}
\usepackage{epstopdf}
\usepackage{url}
\usepackage[square,numbers,sort&compress]{natbib}
\usepackage{amsmath,bm}
\usepackage{stfloats}
\usepackage[pagebackref=false,breaklinks=true,letterpaper=true,colorlinks,bookmarks=false]{hyperref}

\ifCLASSINFOpdf

\else

\fi

\begin{document}

\title{Exploring the Robustness of Human Parsers Towards Common Corruptions}

\author{Sanyi Zhang,~\IEEEmembership{Member,~IEEE,}
        Xiaochun Cao,~\IEEEmembership{Senior Member,~IEEE,}
        Rui Wang,~\IEEEmembership{Member,~IEEE,}\\
        Guo-Jun Qi,~\IEEEmembership{Fellow,~IEEE,}
        and~Jie~Zhou,~\IEEEmembership{Senior Member,~IEEE}
\thanks{

S. Zhang, and R. Wang are with the State Key Laboratory of Information
Security, Institute of Information Engineering, Chinese Academy of
Sciences, Beijing 100093, China,
and also with the School of Cyber Security, University of Chinese Academy of Sciences, Beijing 100049, China (email: zhangsanyi@iie.ac.cn, wangrui@iie.ac.cn).

X. Cao is with the School of Cyber Science and Technology, Shenzhen Campus, Sun Yat-sen University, 518107, China (e-mail: caoxiaochun@mail.sysu.edu.cn).

G.-J. Qi is with Westlake University and OPPO, Hangzhou, 310030, China (e-mail: guojunq@gmail.com).

J. Zhou is with the Department of Automation, State Key Laboratory of Intelligent Technologies and Systems,
Beijing Research Center for Information Science and Technology (BNRist), Tsinghua University, Beijing 100084, China (e-mail:jzhou@tsinghua.edu.cn).
}
}

\markboth{}%
{Shell \MakeLowercase{\textit{et al.}}: Bare Demo of IEEEtran.cls for IEEE Journals}

\maketitle

\begin{abstract}
Human parsing aims to segment each pixel of the human image with fine-grained semantic categories. However, current human parsers trained with clean data are easily confused by numerous image corruptions such as blur and noise. To improve the robustness of human parsers, in this paper, we construct three corruption robustness benchmarks, termed LIP-C, ATR-C, and Pascal-Person-Part-C, to assist us in evaluating the risk tolerance of human parsing models. Inspired by the data augmentation strategy, we propose a novel heterogeneous augmentation-enhanced mechanism to bolster robustness under commonly corrupted conditions. Specifically, two types of data augmentations from different views, \textit{i.e.}, image-aware augmentation and model-aware image-to-image transformation, are integrated in a sequential manner for adapting to unforeseen image corruptions. The image-aware augmentation can enrich the high diversity of training images
with the help of common image operations. The model-aware
augmentation strategy that improves the diversity of input
data by considering the model's randomness. The proposed method is model-agnostic, and it can plug and play into arbitrary state-of-the-art human parsing frameworks.
The experimental results show that the proposed method demonstrates good universality which can improve the robustness of the human parsing models and even the semantic segmentation models when facing various image common corruptions.
Meanwhile, it can still obtain approximate performance on clean data.
\end{abstract}


\section{Introduction}
\IEEEPARstart{H}{uman} parsing is a fine-grained semantic segmentation task that focuses on assigning pixels of the human body regions as semantic categories.
Correctly understanding semantic parts of the human  
can benefit multiple human-related scene understanding tasks, such as fashion landmark detection \cite{wang2018attentive}, virtual try-on \cite{dong2019towards, yang2020towards, ge2021disentangled}, person generation \cite{song2020unpaired, men2020controllable} and person re-identification \cite{zhu2020identity}, human-related scene interaction  \cite{icassp22sgg, acmmm23psg, qi2018learning, zhou2021cascaded, fang2019graspnet}, human body super-resolution \cite{liu2021super}, and so on.

The current human parsing models \cite{li2020self, wang2021hierarchical, ruan2019devil} have achieved superior performance on the challenging datasets collected from real-world scenes. These human parsers are usually trained and tested with clean data. However, the image quality is easily influenced by unexpected distortions in complex real-world scenarios. We hope these well-trained human parsers not only obtain better performance on clean data but also can deal with unrestricted data, \textit{i.e.}, commonly corrupted images. For example, while parsing one person in an outdoor environment under severe weather conditions, or in an indoor environment with bright light, current human parsers may suffer performance drops since the image quality is usually low.
If we apply well-trained human parsers to safety-aware scenarios, the model's ability will be susceptible to image quality.
Thus, human parsers must own good robustness to those possible image corruptions.

\begin{figure}[t]
  \centering
  \includegraphics[width=0.99\columnwidth]{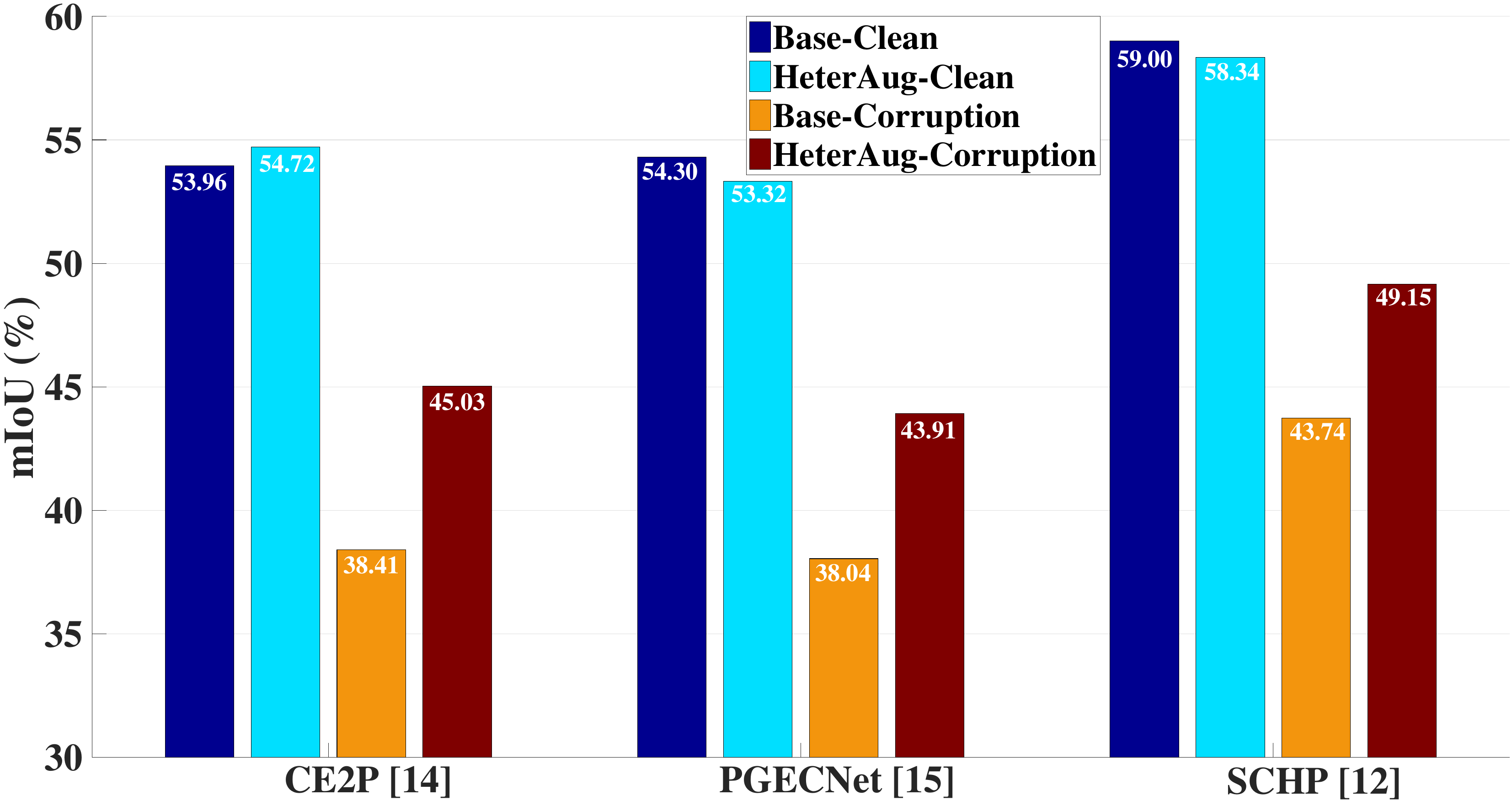}
  \caption{ The improvements of the base model vs. the proposed heterogeneous augmentation (HeterAug) method on corrupted LIP-C dataset. Specifically, three different human parsers, \textit{i.e.}, CE2P \cite{ruan2019devil}, PGECNet \cite{zhang2021human} and SCHP \cite{li2020self}, are conducted. *-Clean and *-Corruption means the methods tested on clean and corrupted validation sets, respectively. The base methods suffer performance drops on the corrupted data, while the HeterAug mechanism obtains approximate performances on the clean data and owns large margins on the corrupted ones.
  }
  \label{fig:first}
\end{figure}

The common robustness has been studied on various tasks, like image classification \cite{hendrycks2019augmix}, object detection \cite{michaelis2019dragon}, semantic segmentation \cite{kamann2020benchmarking,kamann2020increasing}, human pose estimation \cite{wang2021human}, \textit{etc}. In this paper, we mainly focus on exploring the model robustness of the human parsing task. Model robustness is usually defined that a model trained with clean data that should own a powerful ability while being validated on both the clean data and corrupted data.
The main motivation of this paper is to adapt the human parsers to unforeseen input images, and the claim is that they should own good practicability and robustness with respect to various image corruptions while applied to real-world applications.
The key challenges of robust human parsing models are as follows: 1) Current challenging datasets only provide us evaluation metric scores on the clean data. They lack a rigorous benchmark for real-world image corruptions to evaluate the robustness of the state-of-the-art human parsers.
2) The performances of the clean and corrupted images need to be balanced, where an ideal robust human parser should maintain approximate performances in these two cases.
3) Data augmentation-related methods have been certificated to improve the robustness of the model while facing the image corruption problem. However, how
to construct effective augmentation strategies bolstering the generalization of human parsing models with respect to unforeseen image corruption is still an open problem.

To that end, we construct a robustness benchmark for the human parsing task which is implemented on the validation set of current challenge datasets. Specifically, three benchmarks are established, \textit{i.e.}, LIP-C, ATR-C, and Pascal-Person-Part-C, which are constructed on the LIP \cite{liang2018look}, ATR \cite{liang2015deep} and Pascal-Person-Part \cite{chen2014detect} datasets respectively. Through analyzing the evaluation results of the existing human parsing models, there are some observations: 1) All state-of-the-art human parsers suffer different levels of performance plummet on various corrupted images. 2) The current human parsers are easily confused by the Gaussian noise and snow, and they are robust to the brightness and spatter.

As we know, the advantages of data augmentation are to enrich the variety of the training data and enhance the learning difficulty while in the training stage.
Motivated by the data augmentation strategy, we propose a novel heterogeneous augmentation (HeterAug) mechanism to improve the robustness of the human parsers, which enriches the samples through two different views, \textit{i.e.}, image-aware augmentation and model-aware augmentation.
Specifically, image-aware augmentation resorts to the operations of the image library, which first processes the input image via a chain augmentation and then mixes it with the input image.
The other model-aware augmentation perturbs the input samples by the deep neural network, which stochastically resets the parameters of the neural network at each feed-forward step.
Finally, these two different types of data augmentation are joined in a successive manner. The performance of the human parsers with HeterAug can obtain large gains compared with the original base models, as shown in Figure \ref{fig:first}.
Since the data augmentation is just implemented in the training stage and the main human parsing framework is constant, thus the proposed model does not introduce too much extra computation overhead and the inference time remains the same as the original model.
The proposed model is model-agnostic and owns a good ability to deal with various image common corruptions. In addition, extensive experiments have proved that the heterogeneous augmentation mechanism owns good generalization ability which boosts the robustness of the semantic segmentation model on Pascal VOC 2012 dataset \cite{everingham2010pascal}.
The main contributions of this paper are summarized as follows.

\begin{itemize}
    \item[$\diamond$] We construct three novel benchmarks to evaluate the robustness of human parsing models, namely LIP-C, ATR-C, and Pascal-Person-Part-C. The current state-of-the-art human parsers all suffer various levels of performance drop under commonly corrupted conditions. This motivates us attaching importance to the robustness against complex corruptions and encourages us to design suitable models to solve this problem.
    \item[$\diamond$] We propose a novel heterogeneous augmentation (HeterAug) strategy to improve the robustness of the human parsers, which enriches the training data through two different views, \textit{i.e.}, image-aware and model-aware augmentations, and sequentially merges them. The proposed HeterAug mechanism is model-agnostic, allowing us to plug and play it into arbitrary models.
    \item[$\diamond$] The proposed heterogeneous augmentation strategy exhibits good generality, significantly improving the robustness of human parsers and bolstering the ability of the semantic segmentation model against common corruptions.

\end{itemize}

The remaining parts of our paper are as follows. In section \ref{relatework}, we first introduce the related work. Then, the details of the heterogeneous augmentation methodology and the robust human parsing benchmark are presented in Section \ref{method}. Experiments on
three human parsing benchmarks and one semantic segmentation benchmark are discussed in Section \ref{experiment}.
Finally, we conclude in Section \ref{conlusion}.

\begin{figure*}[t]
  \centering
  \includegraphics[width=0.97\linewidth]{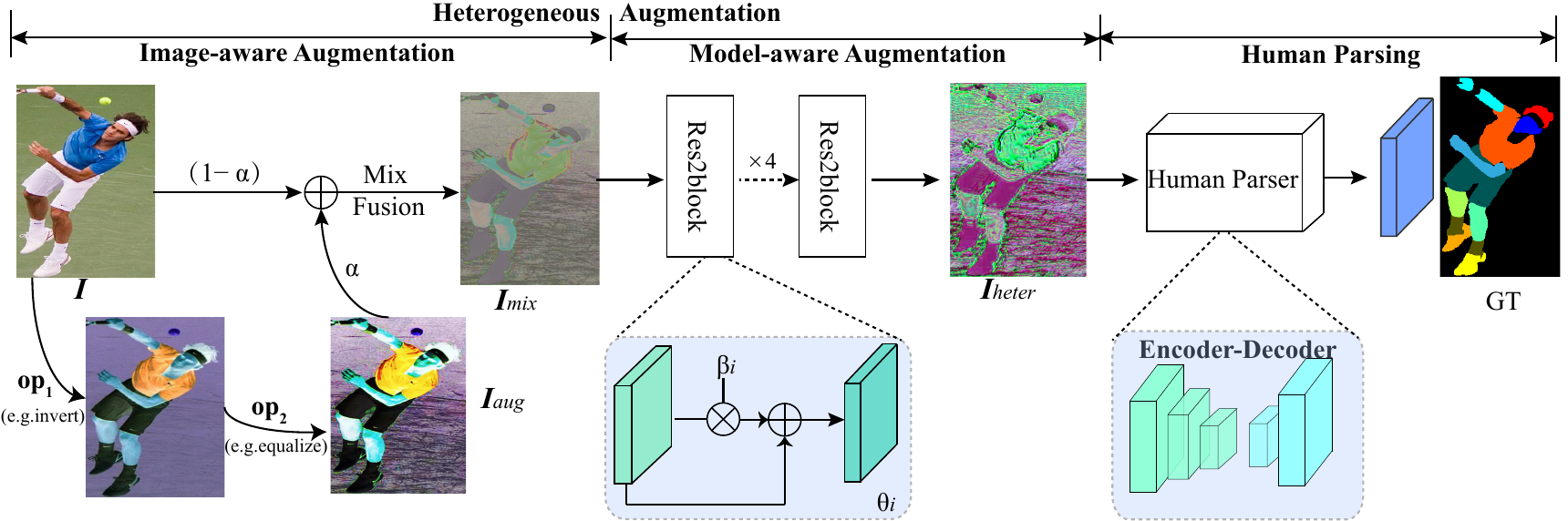}
  \caption{ The framework of the proposed heterogeneous augmentation network. For an input image $I$, it is firstly processed with image-aware augmentation. Specifically, a chain image operations, \textit{i.e.}, $op_{1}$(\textit{e.g.,} invert) and $op_{2}$ (\textit{e.g.,} equalize), are implemented on $I$ to obtain a augmented image $I_{aug}$, then $I$ and $I_{aug}$ are combined with mix operation to obtain the output image $I_{mix}$. The mixed image $I_{mix}$ is further fed into the model-aware augmentation module with random ratios. Finally, the augmented image $I_{heter}$ is employed to train the human parser.
  }
  \label{fig:framework}
\end{figure*}

\section{Related Work}\label{relatework}

In this section, we first introduce current human parsing solutions, and then we will discuss corruption robustness-related solutions in different computer vision tasks.

\subsection{Human Parsing} Human parsing is a specific fine-grained semantic segmentation task that focuses on pixel-level human understanding. Human parsing methods can be categorized into
single-person human parsing and multi-person human parsing. The main difference is that multi-person human parsing needs to locate different human instances. Here we mainly discuss single-person human parsing-related methods.
Current single-person human parsing solutions usually leverage various clues for designing suitable human parsing models.
Early human parsing models devote to developing solutions through hand-crafted cues, for example, low-level image decompositions \cite{yamaguchi2012parsing, liu2013fashion}, human part templates \cite{bo2011shape,dong2013deformable}, human body model via CRF \cite{yamaguchi2013paper} or grammar structure \cite{zhu2008max}. However, the parsing ability of these parsers will be limited by the hand-crafted features.

Recently, researchers focus on exploring solutions via deep learning-oriented designs. Some methods tend to use template matching mechanism \cite{liang2015deep}, combine local and global information \cite{liang2015human}, and gather spatial information via LSTMs \cite{liang2016semantic, liang2016eccv}. To enhance performance, additional inputs such as super-pixels \cite{liang2015human} or HOG feature \cite{luo2013pedestrian} are commonly introduced in these solutions. However, the extra inputs will bring computational burdens. Inspired by the FCN-based architecture in pixel-wise prediction tasks, researchers attempt to draw useful clues from the semantic segmentation models like multi-scale context, human inherent structure and human pose,  pixel semantic relations across different images \cite{wang2021exploring}, non-parametric model based on 
non-learnable prototypes \cite{zhou2022rethinking}, \textit{etc}. Multi-scale context information can effectively solve the various scale problem of different human parts, and it is formulated through the pyramid structure, for example, JPPNet \cite{liang2018look} with ASPP (Atrous Spatial Pyramid Pooling), CE2P \cite{ruan2019devil} with PSP (Pyramid Spatial Parsing), PGECNet \cite{zhang2021human} with PGEC (Pyramidical Gather-Excite Context).
Human part edge information is useful to distinguish adjacent human classes, and it is integrated into the pipeline to improve the parsing boundaries of human parts \cite{ruan2019devil, zhang2021human}.
Human pose information can model the kinetic relation in a connected graph, which is also exploited by human parsers, for instance CorrPM \cite{zhang2020CVPR}, differentiable dense-to-sparse (parsing-to-pose) multi-granularity instance-level human parsing \cite{zhou2021differentiable}. However, these solutions rely on an off-the-shelf human pose estimator or additional human pose annotation, which are time-consuming or not an end-to-end case. Thus, how to explore the human inherent structure without additional annotation or computational burden needs to be further discussed. The human structure is consistent with human understanding, such as visual attention in the underlying rationale behind video object patterns \cite{wang2020paying}. 
To solve these problems, the human inherent structures are explored by utilizing human body hierarchy, \textit{e.g}, as semantic neural tree \cite{ji2019learning}, human compositional structure (CNIF) \cite{wang2019learning}, decomposition and part-relation of hierarchical representation (PRHP) \cite{wang2021hierarchical, wang2020hierarchical}, cross-dataset sharing within graph model \cite{gong2019graphonomy,he2019grapy}, hierarchical tree-shaped relations \cite{li2022deep}. In addition, some useful attempts are also used, such as adversarial training \cite{luo2018macro}, cross-domain knowledge \cite{liu2018cross}, multi-mutual consistency strategy for human motion segmentation \cite{zhou2022consistency}, and self-correction learning (SCHP) \cite{li2020self}, \textit{etc}. Insufficient training samples are limited to training robust human parsers, semi-supervised methods are conducted to solve this problem. For example, a self-learning rectification strategy \cite{li2020selfsemi} equipped with global and local consistencies is proposed.

\subsection{Corruption Robustness}
Current deep learning models are vulnerable to various image corruptions, and researchers have resorted to solutions to improve the robustness of the model.
Corruption robustness has been studied in many computer vision tasks, for example, image classification \cite{hendrycks2019augmix}, object detection \cite{michaelis2019dragon}, semantic segmentation \cite{kamann2020benchmarking,kamann2020increasing}, human pose estimation \cite{wang2021human}, and so on. The core challenge is how to improve the model's robustness to unforeseen data shifts.
To solve the robustness problem, researchers have developed various mechanisms, including data augmentation \cite{wang2021augmax}, pre-training \cite{jiang2020robust}, Lipschitz continuity \cite{weng2018evaluating}, stability training \cite{zheng2016improving}, DeepAugment \cite{hendrycks2021many} and so on. Specifically, data augmentation is the widely explored one.

For image classification benchmark, AugMix \cite{hendrycks2019augmix} mixes multiple randomly augmented images as the data augmentation strategy
and employs a Jensen-Shannon loss enforcing consistency among diverse inputs to improve the model's robustness. AugMax \cite{wang2021augmax} considers both diversity and hardness 
in training by introducing a new approach that combines various augmentation operators in a random mixture to create a more challenging training experience.
AugMax \cite{wang2021augmax} designs a disentangled
normalization mechanism named DuBIN to solve the
high heterogeneity of input distribution for achieving a better
robustness on the image classification task.

The human pose estimation benchmark is a multi-task prediction that contains human body keypoint classification and regression. AdvMix \cite{wang2021human} proposes a novel adversarial data augmentation with two neural networks and knowledge distillation to maintain the performance of the clean data.

For object detection benchmark, Michaelis \textit{et al.} \cite{michaelis2019dragon} improve the robustness through stylizing the training images.

For the semantic segmentation task, Kamann \textit{et al.} \cite{kamann2020benchmarking, kamann2021benchmarking} provide a detailed benchmark study and robust model design rules for the semantic segmentation models. Kamann \textit{et al.} \cite{kamann2020increasing} propose a novel Painting-by-Numbers data augmentation technology to enhance the model robustness. In addition, Xu \textit{et al.} \cite{xu2021dynamic} propose a divide-and-conquer adversarial training mechanism for solving adversarial attack problems in the semantic segmentation task.

\section{Methodology}\label{method}

\subsection{Heterogeneous Augmentation}

The proposed heterogeneous augmentation framework is shown in Figure \ref{fig:framework}. Given one input image $I$, two types of image augmentation, \textit{i.e.}, image-aware and model-aware corrupted augmentations, are sequentially employed to enhance the diversity of the training samples. In particular, image-aware augmentation is implemented through a chain image augmentation and mixed with the input image. And the model-aware augmentation is achieved via the deep neural network with stochastic weights.

\textbf{Image-aware Augmentation.}
The goal of image-aware augmentation is to enrich the high diversity of training images with the help of the image operation library which can improve the model's robustness. Motivated by AutoAugment \cite{cubuk2018autoaugment}, a chain of image operations upon a well-searched augmentation policy is utilized. For input image $I$, we sample $k$ operations from a pre-defined image operation set $\mathcal{O} = \left\{\text{op}_{1}, ..., \text{op}_{K}\right\}$.
Specifically, $k$ image operations\footnote{All image operations can be easily achieved via the PIL library.} are sequentially performed on the image, and each operation is implemented with a constant magnitude. To raise the randomness of image-aware augmentation, each operation is implemented with a random probability. The augmented image $I_{aug}$ can be denoted as:

\begin{equation}
I_{aug} = chain(I), chain = \text{op}_{1} \circ \text{op}_{2},..., \text{op}_{k}.
\end{equation}

Then, we combine the original image $I$ and the processed image $I_{aug}$ with the mix operation.

\begin{equation}
I_{mix} = \alpha I + (1-\alpha) I_{aug},
\end{equation}
where $\alpha$ is sampled from a Beta distribution.

\textbf{Model-aware Augmentation.} 
The image-aware augmentation utilizes the standard image operations making the model adapts to unforeseen image corruption. However, image corruptions usually contain various variants in real-world scenarios.
Current state-of-the-art solutions for human parsing tasks resort to the deep neural network for good recognition abilities.
Thus, we further explore the model-aware augmentation strategy that improves the diversity of input data by taking the model randomness into account.
In light of the pixel-wise segmentation goal, we employ an image-to-image pipeline to keep the spatial resolution of output consistent with the input image.
The deep neural network is randomly sampled according to the pre-designed structure. The weights of the network are randomly generated along with different iterations. In particular, the deep augment network $F$ is constructed with a series of residual blocks $B_{i}$, and we also introduce a control parameter $\beta$ to further enhance the randomness of the sample. The size of input and output samples are consistent. For each block $B_{i}$, the weight of block $B_{i}$ is denoted as $\theta_{i}$ and the control parameter is set as $\beta_{i}$. The output feature of $i$-th block is denoted as:

\begin{equation}
B_i(x) = x + \beta_{i} F_{i}(x, \theta_{i}).
\end{equation}

The basic residual block adopts the Res2block \cite{gao2019res2net}. The main framework is randomly generated and initialized at the beginning, and the weights of the deep augment network are randomly re-sampled at each mini-batch. The control parameter $\beta_{i}$ is also randomly sampled for each residual block at every mini-batch.

Finally, the image-aware and model-aware augmentation strategies are implemented successively. Thus, the output image of the deep augmentation network $I_{heter}$ is denoted as:

\begin{equation}
I_{heter} = F(I_{mix}; \theta, \beta),
\end{equation}
where $\theta$ and $\beta$ are the parameters of the deep network $F$.

\textbf{Human Parser.}
While the input images are preprocessed by the heterogeneous augmentation policy, these images will be fed into the following human parsing model. The human parser is the standard fully convolutional neural network (FCN) equipped with the encoder-decoder framework. The whole human parsing model is trained by optimizing the pixel-wise cross-entropy loss and other auxiliary losses.

\begin{figure}[!th]
  \centering
  \includegraphics[width=0.85\linewidth]{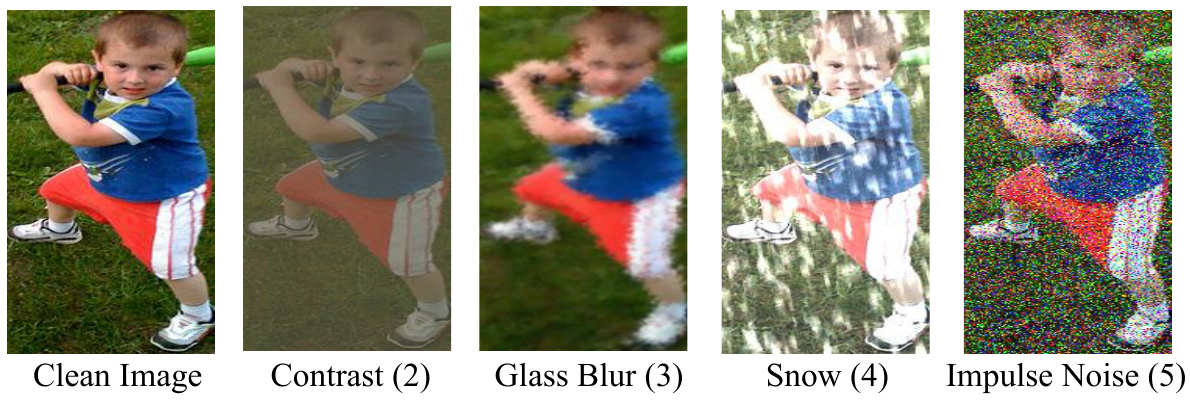}
  \caption{Visualization of example images in the robust human parsing benchmark LIP-C dataset. We sample one corruption type (severity level 1 to 5) from blur, digital, noise and weather categories, respectively.
  }
  \label{fig:example}
\end{figure}

\begin{table*}
\centering
\footnotesize
\caption{
Average clean and corrupted mIoU comparison results on LIP-C dataset.
 }
\label{tab:lip}
\small{
\resizebox{0.85\textwidth}{!}{
\begin{tabular}{ p{3.2cm} | c | c c c c | c c c c }
\toprule
\multirow{2}{*}{Method}  & \multirow{2}{*}{Clean} & \multicolumn{4}{c}{Blur }& \multicolumn{4}{|c}{Digital} \\
\cline{3-10}
~& ~& Defocus & Gaussian & Motion & Glass & Brightness & Contrast & Saturate & JPEG \\
\hline
JPPNet \cite{liang2018look}              & 51.37  & 36.33 & 37.95 & 36.45 & 35.97 & 45.46 & 31.08 & 44.94 & 42.09   \\
MMAN \cite{luo2018macro} & 47.66  & 35.87 & 37.11 & 37.46 & 37.60 & 44.94 & 42.21 & 43.35 & 43.53    \\
CE2P \cite{ruan2019devil}               & 53.96  & 37.63 & 39.33 & 36.58 & 37.75 & 48.96 & 31.62 & 47.58 & 45.50     \\
CorrPM \cite{zhang2020CVPR}             & 55.33  & 40.00 & 41.62 & 38.67 & 39.55 & 50.20 & 35.37 & 48.93 & 48.19     \\
PGECNet \cite{zhang2021human}             & 54.30  & 37.83 & 39.52 & 37.36 & 38.44 & 49.14 & 33.11 & 48.13 & 44.92     \\
HRNet-W48  \cite{WangSCJDZLMTWLX19}         & 55.83  & 39.16 & 41.01 & 38.31 & 37.45 & 51.15 & 36.51 & 50.03 & 46.26 \\
HRNet-W48-OCR \cite{YuanCW19}     & 56.48  & 39.91 & 41.89 & 38.94 & 38.74 & 51.97 & 36.95 & 51.02 & 47.90 \\
SCHP \cite{li2020self}              & \textbf{59.00} & 44.08 & 46.05 & 43.05 & 43.51 & 54.61 & 39.30 & 53.45 & 51.32 \\
\midrule
PGECNet-HeterAug & 53.32 & 39.62 & 41.02 & 38.44 & 38.59 & 52.51 & 45.86 & 51.25 & 48.80 \\
CE2P-HeterAug & 54.72 & 40.76 & 42.28 & 40.02 & 40.42 & 52.99 & 46.63 & 51.85 & 49.48 \\
SCHP-HeterAug                                & 58.34 & \textbf{45.79} & \textbf{47.35} & \textbf{43.93} & \textbf{46.02} & \textbf{56.52} & \textbf{50.34} & \textbf{55.68} & \textbf{53.08} \\
\bottomrule
\end{tabular}
}\footnotesize

\resizebox{0.85\textwidth}{!}{
\begin{tabular}{  p{3.05cm} | c c c c | c c c c | c}

\toprule
\multirow{2}{*}{Method}  & \multicolumn{4}{c}{Noise}& \multicolumn{4}{|c|}{Weather} & \multirow{2}{*}{mIoU$_{c}$} \\
\cline{2-9}
~ &  Gaussian & Impulse & Shot & Speckle & Fog & Frost & Snow & Spatter & ~ \\
\hline
JPPNet \cite{liang2018look}              & 28.42 & 26.39 & 28.73 & 35.33 & 41.10 & 36.86 & 31.59 & 40.98 & 36.23 \\
MMAN \cite{luo2018macro}                & 34.65 & 33.45 & 35.16 & 39.26 & 39.44 & 36.61 & 35.73 & 41.08 & 38.59 \\
CE2P \cite{ruan2019devil}                & 32.07 & 32.03 & 32.87 & 39.62 & 41.15 & 37.78 & 31.02 & 43.06 & 38.41 \\
CorrPM \cite{zhang2020CVPR}                 & 29.73 & 28.04 & 30.79 & 37.98 & 43.91 & 39.47 & 33.00 & 43.81 & 39.33 \\
PGECNet \cite{zhang2021human}            & 28.83 & 27.80 & 30.10 & 38.51 & 42.01 & 37.55 & 31.00 & 44.37 & 38.04 \\
HRNet-W48 \cite{WangSCJDZLMTWLX19}          & 30.64 & 31.09 & 32.00 & 39.86 & 45.45 & 41.16 & 35.41 & 45.99 & 40.09 \\
HRNet-W48-OCR \cite{YuanCW19}      & 30.54 & 30.78 & 32.06 & 41.23 & 45.86 & 41.30 & 35.15 & 47.66 & 40.74 \\
SCHP \cite{li2020self}             & 33.02 & 30.44 & 34.71 & 42.96 & 49.49 & 45.39 & 39.41 & 49.07 & 43.74 \\
\midrule
PGECNet-HeterAug & 40.89 & 40.42 & 42.38 & 46.27 & 49.01 & 43.33 & 36.36 & 47.88 & 43.91 \\
CE2P-HeterAug                           & 42.40 & 42.16 & 43.56 & 46.98 & 49.38 & 43.68 & 39.46 & 48.50 & 45.03 \\
SCHP-HeterAug                              & \textbf{46.06} & \textbf{45.10} & \textbf{47.38} & \textbf{50.97} & \textbf{53.93} & \textbf{48.82} & \textbf{43.08} & \textbf{52.41} & \textbf{49.15} \\

\bottomrule
\end{tabular}
}
}
\end{table*}

\subsection{Robust Human Parsing Benchmark}

\textbf{Benchmark Datasets.}
To evaluate the robustness of the human parsing models, we construct a robust human parsing benchmark implemented on the validation set of standard human parsing datasets, which contains 16 different types of image corruptions. Each image corruption employs 5 severity levels. All corruption types can be categorized into four groups, \textit{i.e.}, blur (defocus, gaussian, motion, glass), noise (gaussian, impulse, shot, speckle), digital (brightness, contrast, saturate, JPEG compression), and weather (fog, frost, snow, spatter). The common corruptions are implemented through the python \textit{imagecorruptions} package\footnote{\url{https://github.com/bethgelab/imagecorruptions}}.Three robust human parsing benchmarks, \textit{i.e.}, LIP-C, ATR-C, and Pascal-Person-Part-C, are constructed on the LIP \cite{liang2018look}, ATR \cite{liang2015deep}, and Pascal-Person-Part \cite{chen2014detect} datasets, respectively.

\textbf{LIP-C dataset} \cite{liang2018look} is constructed on the LIP validation set which contains $10,000$ images. $16$ image corruption types are conducted on the validation set. The total number of LIP-C is $16 \times 5$ times that of the clean LIP val set. The standard LIP training set contains $30,462$ images, and there are $20$ classes (19 semantic categories, and 1 background class).

\textbf{ATR-C dataset} \cite{liang2015deep} is constructed on the ATR testing set which contains $1,000$ images. In the standard ATR dataset, there are $17,700$ images in total. $16,000$ and $700$ images are used for training and validation, respectively.
Each image is annotated with $18$ fine-grained semantic classes.

\textbf{Pascal-Person-Part-C dataset} \cite{chen2014detect} is constructed on the Pascal-Person-Part validation set which contains $1,817$ images. Pascal-Person-Part is a relatively small-scale dataset that is a subset of PASCAL VOC 2010, with only six semantic parts and one background class. There are $1,716$ images in the training set.

In Figure \ref{fig:example}, we show some examples from the LIP-C dataset. There are four different categories in the robustness benchmark, \textit{i.e.}, blur, digital, noise, and weather. Specifically, we sample one image corruption from each category, \textit{i.e.}, glass blur, contrast, impulse noise, and snow. Each image corruption operation is implemented with a random severity level. If a large severity level is applied, the image will be severely destroyed. The large severity level means that the difficulty of human parsing will increase.

\begin{table*}[t]
\centering
\footnotesize
\caption{Average clean and corrupted mIoU comparison results on ATR-C dataset. }
\resizebox{0.85\textwidth}{!}{
\begin{tabular}{ p{2.5cm} | c | c c c c | c c c c }
\toprule
\multirow{2}{*}{Method}  & \multirow{2}{*}{Clean} & \multicolumn{4}{c}{Blur}& \multicolumn{4}{|c}{Digital} \\
\cline{3-10}
~& ~& Defocus & Gaussian & Motion & Glass & Brightness & Contrast & Saturate & JPEG \\
\hline
TGPNet \cite{Luo2018TGPnet}         & 73.46 & 39.81  & 42.00 & 53.37 & 56.88 & 68.93 & 51.30 & 63.54 & 69.59  \\
PGECNet \cite{zhang2021human}       & 76.95 & 33.33  & 36.94 & 41.98 & 45.30 & 72.72 & 49.52 & 69.52 & 69.94  \\
Grapy-ML \cite{he2019grapy}          & 77.32 & 41.10 & 44.47 & 48.81 & 50.21 & 73.05 & 51.64 & 70.40 & 69.19 \\
SCHP \cite{li2020self}              &  \textbf{80.65}  & 44.78 & 45.87  & 51.15 & 56.77 & \textbf{77.34} & 60.13 & 75.38 & \textbf{75.50} \\
\midrule
Grapy-ML-HeterAug          & 76.52 & 47.02 & 49.82 & 52.68 & 53.04 & 74.15 & 69.23 & 72.83 & 69.80 \\
SCHP-HeterAug                              &  78.16  & \textbf{56.56} & \textbf{57.40} & \textbf{54.25} & \textbf{57.54} & 76.39 & \textbf{71.27} & \textbf{75.65} & 73.58 \\
\bottomrule
\end{tabular}
}\footnotesize

\resizebox{0.85\textwidth}{!}{
\begin{tabular}{ p{2.5cm} | c c c c | c c c c | c }
\toprule
\multirow{2}{*}{Method}  & \multicolumn{4}{c}{Noise}& \multicolumn{4}{c|}{Weather} & \multirow{2}{*}{mIoU$_{c}$} \\
\cline{2-9}
~ &  Gaussian & Impulse & Shot & Speckle & Fog & Frost & Snow & Spatter & ~ \\
\hline
TGPNet \cite{Luo2018TGPnet}             & 47.12  & 46.69 & 46.35 & 51.41 & 64.06 & 49.09 & 47.64 & 62.21 & 53.75     \\
PGECNet \cite{zhang2021human}        & 32.28  & 33.56 & 30.50 & 40.75 & 68.09 & 54.47 & 49.18 & 67.27 & 49.72 \\
Grapy-ML \cite{he2019grapy}          & 35.34 & 38.51 & 33.38 & 44.05 & 69.24 & 60.11 & 53.20 & 69.18 & 53.24 \\
SCHP \cite{li2020self}               & 50.98  & 51.09 & 49.98 & 57.91 & 75.20 & 67.08 & 61.94 & \textbf{73.56} & 60.92 \\
\midrule
Grapy-ML-HeterAug          & 58.80  & 59.13  & 58.88  & 63.05  & 73.63  & 65.91  & 59.90  & 69.91  & 62.36 \\
SCHP-HeterAug                                 & \textbf{66.14}  & \textbf{66.31} & \textbf{65.56} & \textbf{68.09} & \textbf{76.67} & \textbf{67.90} & \textbf{64.29} & 72.79 & \textbf{66.90} \\
\bottomrule
\end{tabular}
}

\label{tab:atr}
\end{table*}

\textbf{Evaluation Metrics.}
We use the metric score mIoU$_{c}$ to evaluate the robustness of the human parsing models.

\begin{equation}
\text{mIoU}_{\text{c}}=\frac{1}{N_{c}} \sum_{i=1}^{N_{c}} \frac{1}{N_{s}} \sum_{j=1}^{N_{s}} \text{mIoU}_{c}^{i,j},
\end{equation}
where $N_{c}$ and $N_{s}$ denote the number of image corruption types and the severity levels, respectively. In experiments, we set $N_{c}=16$ and $N_{s}=5$.

For the ATR dataset, previous methods usually report the F1 score for comparison. Thus, we also report the performance of F1$_{c}$ score on the ATR-C dataset.
The metric score F1$_{c}$ is defined as:

\begin{equation}
\text{F1}_{\text{c}}=\frac{1}{N_{c}} \sum_{i=1}^{N_{c}} \frac{1}{N_{s}} \sum_{j=1}^{N_{s}} \text{F1}_{c}^{i,j},
\end{equation}
where $N_{c}$ and $N_{s}$ denote the number of image corruption types and the severity levels, respectively.

\section{Experiment}\label{experiment}

\subsection{Experimental Setting}
We evaluate the performance of the state-of-the-art human parsing models on the proposed robust human parsing benchmarks. For a fair comparison, all models are trained on the clean data and tested on the clean and corrupted samples.
We use the publicly well-trained models provided by the authors to test the robustness of models on the corrupted validation sets. 
All comparison models are trained on the standard LIP-training, ATR-training, and Pascal-Person-Part-training sets respectively, and tested on the corrupted validation sets, LIP-C, ATR-C, and Pascal-Person-Part-C, respectively.
In addition, we also provide comparisons on the general semantic segmentation dataset, \textit{i.e.}, Pascal VOC 2012 validation set \cite{everingham2010pascal}.

\begin{table*}[t]
\footnotesize
\centering
\caption{Average clean and corrupted F1 score comparison results on ATR-C dataset. }
\resizebox{0.85\textwidth}{!}{
\begin{tabular}{ p{2.8cm} | c | c c c c | c c c c }
\toprule
\multirow{2}{*}{Method}  & \multirow{2}{*}{Clean} & \multicolumn{4}{c}{Blur}& \multicolumn{4}{|c}{Digital} \\
\cline{3-10}
~& ~& Defocus & Gaussian & Motion & Glass & Brightness & Contrast & Saturate & JPEG \\
\hline

TGPNet \cite{Luo2018TGPnet}               & 83.61  & 49.66 & 51.13 & 66.16 & 69.65 & 80.18 & 62.92 & 76.20 & 80.87     \\
PGECNet \cite{zhang2021human}  & 86.45 & 42.52 & 45.88 & 53.96 & 57.03 & 83.48 & 61.13 & 81.24 & 81.40 \\
Grapy-ML \cite{he2019grapy}          & 86.60 & 51.07 & 54.12 & 61.76 & 62.35 & 83.53 & 62.98 & 81.80 & 80.73 \\
SCHP \cite{li2020self} & \textbf{89.00} & 56.01 & 55.87 & 64.34 & 68.87 & \textbf{86.88} & 72.11 & 85.59 & \textbf{85.55} \\
\midrule
Grapy-ML-HeterAug         & 86.12 & 57.77 & 59.74 & 65.91 & 65.31 & 84.50 & 80.82 & 83.60 & 81.26 \\
SCHP-HeterAug & 87.39  & \textbf{68.89} & \textbf{68.58} & \textbf{67.44} & \textbf{69.84} & 86.20 & \textbf{82.43} & \textbf{85.71} & 84.22 \\
\bottomrule
\end{tabular}
}\footnotesize

\resizebox{0.85\textwidth}{!}{
\begin{tabular}{ p{2.55cm} | c  c c c |c  c c c | c }
\toprule
\multirow{2}{*}{Method}  & \multicolumn{4}{c}{Noise}& \multicolumn{4}{|c|}{Weather} & \multirow{2}{*}{F1$_{c}$} \\
\cline{2-9}
~ &  Gaussian & Impulse & Shot & Speckle & Fog & Frost & Snow & Spatter & ~ \\
\hline

TGPNet \cite{Luo2018TGPnet}              & 60.17 & 60.11 & 59.33 & 64.69 & 75.47 & 63.36 & 61.82 & 75.02 & 66.05 \\
PGECNet \cite{zhang2021human} & 42.14 & 43.52 & 40.26 & 53.00 & 80.16 & 68.78 & 63.36 & 79.31 & 61.07 \\
Grapy-ML \cite{he2019grapy}          &  45.03 & 48.44 & 43.22 & 56.63 & 80.82 & 73.75 & 67.13 & 80.73 & 64.63 \\
SCHP \cite{li2020self} & 62.93 & 62.89 & 62.37 & 70.82 & 85.46 & 79.78 & 75.46 & \textbf{84.27} & 72.45 \\
\midrule
Grapy-ML-HeterAug          & 72.30 & 72.66 & 72.55 & 76.26 & 84.15 & 78.64 & 73.58 & 81.44 & 74.41 \\
SCHP-HeterAug & \textbf{78.64} & \textbf{78.80} & \textbf{78.21} & \textbf{80.29} & \textbf{86.42} & \textbf{80.33} & \textbf{77.30} & 83.71 & \textbf{78.55} \\
\bottomrule
\end{tabular}
}

\label{tab:atrf1}
\end{table*}

\begin{table*}[t]
\centering
\footnotesize
\caption{Average clean and corrupted mIoU comparison results on Pascal-Person-Part-C dataset. }
\resizebox{0.85\textwidth}{!}{
\begin{tabular}{ p{2.55cm} | c | c c c c | c c c c }
\toprule
\multirow{2}{*}{Method}  & \multirow{2}{*}{Clean} & \multicolumn{4}{c}{Blur}& \multicolumn{4}{|c}{Digital} \\
\cline{3-10}
~& ~& Defocus & Gaussian & Motion & Glass & Brightness & Contrast & Saturate & JPEG \\
\hline
Graphomy \cite{gong2019graphonomy}  & 69.15 & 52.88  & 54.09 & 53.06 & 49.05 & 65.46 & 50.85 & 63.54 & 59.13  \\
Grapy-ML \cite{he2019grapy}          & 69.50 & 51.23  & 52.57 & 51.23 & \textbf{52.76} & 66.42 & 49.40 & 64.82 & 56.89 \\
SCHP \cite{li2020self}              & \textbf{70.87} & 47.22  & 49.12 & 53.97 & 50.20 & \textbf{67.06} & 52.36 & \textbf{66.03} & 57.18 \\
\midrule
Grapy-ML-HeterAug & 68.22 & 50.11 & 51.96 & 51.76 & 49.24 & 67.02 & 59.96 & 66.07 & 61.92 \\
SCHP-HeterAug                                & 68.18 & \textbf{53.82}  & \textbf{54.99} & \textbf{54.19} & 48.99 & 66.14 & \textbf{57.98} & 65.53 & \textbf{59.96} \\
\midrule
\end{tabular}
}\footnotesize

\resizebox{0.85\textwidth}{!}{
\begin{tabular}{ p{2.45cm} | c c c c | c c c c | c }
\toprule
\multirow{2}{*}{Method}  & \multicolumn{4}{c}{Noise}& \multicolumn{4}{|c|}{Weather} & \multirow{2}{*}{mIoU$_{c}$} \\
\cline{2-9}
~ &  Gaussian & Impulse & Shot & Speckle & Fog & Frost & Snow & Spatter & ~ \\
\hline
Graphomy \cite{gong2019graphonomy}   & 36.80  & 37.02 & 37.96 & 49.17 & 57.48 & 47.02 & 41.09 & 55.20 & 50.61     \\
Grapy-ML \cite{he2019grapy}           & 38.24  & 39.08 & 40.54 & 52.71 & 59.90 & 50.24 & 42.75 & \textbf{60.16} & 51.81 \\
SCHP \cite{li2020self}               & 44.56  & 43.40 & 46.49 & 55.83 & \textbf{62.33} & 52.12 & 45.83 & 56.90 & 53.16 \\
\midrule
Grapy-ML-HeterAug & 54.09 & 54.07 & 55.79 & 60.11 & 61.63 & 55.05 & 46.59 & 61.48 & \textbf{56.68}  \\
SCHP-HeterAug                                 & \textbf{51.70}  & \textbf{51.19} & \textbf{53.55} & \textbf{58.82} & 61.17 & \textbf{54.02} & \textbf{48.31} & 59.06 & 56.21 \\
\bottomrule
\end{tabular}
}

\label{tab:pascal}
\end{table*}

\subsection{Implementation Details}

As suggested in \cite{cubuk2018autoaugment}, for the image-aware augmentation step, we sample two image operations for each image. And the image operation set contains \textit{equalize, posterize, solarize, invert, and sharpness}, which has no overlap with the tested image corruptions.
For the model-aware augmentation step, four successive Res2Net modules are used. The range of the random control parameter $\beta$ is set as $[0.375, 0.75]$.
For the human parser training, we follow the same training strategy suggested in the state-of-the-art human parsing models, such as the initial learning rate (\textit{e.g.}, 0.007), the optimizer (\textit{e.g.}, Poly), the training epochs (\textit{e.g.}, 150), and so on. All experiments are conducted on PyTorch on NVIDIA TITAN RTX GPUs. All evaluation results are achieved with the publicly well-trained models by the authors.
The detail GitHub links of state-of-the-art human parsers are JPPNet\footnote{\url{https://github.com/Engineering-Course/LIP\_JPPNet}},
MMAN\footnote{\url{https://github.com/RoyalVane/MMAN}},
CE2P\footnote{\url{https://github.com/liutinglt/CE2P}},
PGECNet\footnote{\url{https://github.com/31sy/PGECNet}},
CorrPM\footnote{\url{https://github.com/ziwei-zh/CorrPM}},
HRNet-W48 and HRNet-W48-OCR\footnote{\url{https://github.com/HRNet/HRNet-Semantic-Segmentation}},
SCHP\footnote{\url{https://github.com/PeikeLi/Self-Correction-Human-Parsing}},
TGPNet\footnote{\url{https://github.com/curious999/TGPnet}},
Graphonomy\footnote{\url{https://github.com/Gaoyiminggithub/Graphonomy}},
and Grapy-ML\footnote{\url{https://github.com/Charleshhy/Grapy-ML}}.
We implement the proposed heterogeneous augmentation strategy on two different human parsers, \textit{i.e.}, SCHP\footnote{\url{https://github.com/PeikeLi/Self-Correction-Human-Parsing}} \cite{li2020self}
and CE2P\footnote{\url{https://github.com/liutinglt/CE2P}} \cite{ruan2019devil} to verify its effectiveness. For the semantic segmentation task, we use Deeplabv3plus with ResNet-101 backbone to verify its effectiveness.

\subsection{Quantitative Results}

\textbf{Comparison with state-of-the-arts.}
In Table \ref{tab:lip}, Table \ref{tab:atr}, Table \ref{tab:atrf1} and Table \ref{tab:pascal}, we report the evaluation scores comparing with state-of-the-art human parsing models on the LIP-C, ATR-C, Pascal-Person-Part-C datasets, respectively.
It can be observed that the SCHP model with our proposed heterogeneous augmentation (SCHP-HeterAug) strategy outperforms other state-of-the-art methods by a large margin.

As shown in Table \ref{tab:lip}, all human parsing models on LIP-C are usually robust to the brightness and saturation where the performance drops are smaller than other image corruptions. But human parsers are not good at the image contrast condition. While we employ heterogeneous data augmentation, the performance of image contrast can obtain a big promotion. Meanwhile, almost all human parsing models are very sensitive to different image noises and these methods suffer the performances plummet. Generally, the standard image augmentation strategy in almost all current human parsing model training mainly adopts spatial transformations, like image scaling, horizontal flipping, and so on. The tested corruptions are focused on the non-spatial transformations, thus the current human parsers cannot handle the image noises well. In particular, our proposed HeterAug mechanism boosts performance on all noise-related corruptions.
For the weather corruptions, current parsing models are more robust to fog and spatter, and less robust to frost and snow conditions. The proposed heterogeneous augmentation method can achieve robust gains across all image corruptions.
Specifically, SCHP-HeterAug obtains $5.41\%$ gains than SCHP, and the performances on clean data are close. The HeterAug mechanism also demonstrates its effectiveness on other human parsers, \textit{i.e.}, CE2P and PGECNet. These good abilities show the robustness of our proposed model.

\begin{table}[t]
\small
		\caption{Different random ratios of chain augmentation in image-aware augmentation on LIP-C dataset.}
		\label{tab:chain}
		\centering
\begin{tabular}{c | c c  }
\hline
Random Ratio & Clean & mIoU$_{c}$  \\
\hline
0\% & 53.96 & 38.41 \\
30\% & 55.05 & 42.58 \\
50\% & 55.03 & 43.00 \\
70\% & \textbf{55.34} & 43.13 \\
100\% & 55.11 & \textbf{43.25} \\
\hline
\end{tabular}

\end{table}

\begin{table}
\small
\centering
		\caption{Different ratios of model-aware augmentation on LIP-C dataset.}
		\label{tab:deepaug}
\begin{tabular}{c | c  c  }
\hline
Random Ratio & Clean & mIoU$_{c}$  \\
\hline
0\% & 53.96 & 38.41 \\
25\% & 54.90 & \textbf{43.70} \\
50\% & 51.41 & 38.39 \\
70\% & \textbf{55.39} & 40.62 \\
90\% & 52.51 & 42.04 \\
\hline
\end{tabular}
\end{table}

\begin{table}
\small
\centering
\caption{Different composition types comparison of heterogeneous augmentation on LIP-C dataset.}
\label{tab:fuse}
\begin{tabular}{c | c  c  }
\hline
Method & Clean & mIoU$_{c}$  \\
\specialrule{0em}{1pt}{1pt}
\hline
\specialrule{0em}{1.4pt}{1.4pt}
CE2P & 53.96 & 38.41 \\
\specialrule{0em}{0.5pt}{0.5pt}
Random  & 54.48 & 43.90 \\
\specialrule{0em}{1pt}{1pt}
Seq & \textbf{54.67} & \textbf{44.48} \\
\specialrule{0em}{1pt}{1pt}
\hline
\end{tabular}

\end{table}

\begin{table*}[th]
\centering
\footnotesize
\caption{The effectiveness of different augmentation mechanisms on LIP-C dataset.}
\label{tab:mix}
\resizebox{0.85\textwidth}{!}{
\begin{tabular}{ p{2.4cm} | c | c c c c | c c c c }
\toprule
\multirow{2}{*}{Method}  & \multirow{2}{*}{Clean} & \multicolumn{4}{c}{Blur}& \multicolumn{4}{|c}{Digital} \\
\cline{3-10}
~& ~& Defocus & Gaussian & Motion & Glass & Brightness & Contrast & Saturate & JPEG \\
\hline
CE2P \cite{ruan2019devil}               & 53.96  & 37.63 & 39.33 & 36.58 & 37.75 & 48.96 & 31.62 & 47.58 & 45.50     \\
\midrule
+Grid Mask \cite{chen2020gridmask} & 55.00 & 40.50 & 42.09 & 39.03 & 39.42 & 50.07 & 35.33 & 48.83 & 47.92 \\
+ImageAug & 55.31 & 40.66 & 42.09 & 38.96 & 39.02 & 53.02 & 46.29 & 51.12 & 49.21 \\
+ModelAug  & 54.90 & 40.89 & 42.59 & 40.23 & 40.47 & 52.31 & 38.08 & 51.87 & 49.66 \\
+HeterAug & 54.72 & 40.76 & 42.28 & 40.02 & 40.42 & 52.99 & 46.63 & 51.85 & 49.48 \\

\bottomrule
\end{tabular}
}\footnotesize

\resizebox{0.85\textwidth}{!}{
\begin{tabular}{ p{2.25cm} | c  c c c |c  c c c | c }
\toprule
\multirow{2}{*}{Method}  & \multicolumn{4}{c}{Noise}& \multicolumn{4}{|c|}{Weather} & \multirow{2}{*}{mIoU$_{c}$} \\
\cline{2-9}
~ &  Gaussian & Impulse & Shot & Speckle & Fog & Frost & Snow & Spatter & ~ \\
\hline
CE2P \cite{ruan2019devil}                & 32.07 & 32.03 & 32.87 & 39.62 & 41.15 & 37.78 & 31.02 & 43.06 & 38.41 \\
\midrule
+Grid Mask \cite{chen2020gridmask}  & 32.77 & 32.51 & 34.46 & 40.96 & 44.42 & 41.21 & 36.16 & 45.78 & 40.72 \\
+ImageAug  & 39.44 & 39.18 & 40.65 & 45.50 & 49.36 & 42.15 & 36.10 & 48.23 & 43.81 \\
+ModelAug & 40.76 & 40.06 & 41.51 & 45.23 & 46.52 & 42.79 & 38.88 & 47.36 & 43.70 \\
+HeterAug & 42.40 & 42.16 & 43.56 & 46.98 & 49.38 & 43.68 & 39.46 & 48.50 & 45.03 \\

\bottomrule
\end{tabular}
}

\end{table*}

The quantitative results of ATR-C and Pascal-Person-Part-C in Table \ref{tab:atr}, Table \ref{tab:atrf1} and Table \ref{tab:pascal} also have similar observations with the LIP-C dataset. The proposed heterogeneous augmentation method can achieve good robustness with large improvements over other state-of-the-art human parsing methods.

\subsection{Ablation Studies}

To certify the effectiveness of the proposed heterogeneous augmentation strategy, we conduct ablation experiments on the LIP-C dataset and choose the CE2P model as the baseline. For a fair comparison, we follow the same training setting with CE2P and use horizontal flipping while testing.

\textbf{The effectiveness of chain augmentation in image-aware augmentation.} We randomly conduct the chain augmentation policy with different ratios to verify the model's robustness. Specifically, 0\%, 30\%, 50\%, 70\%, 100\% of training images are pre-processed with chain augmentation in the training stage, as shown in Table \ref{tab:chain}. If we randomly process the training images with chain augmentation, the performance of LIP-C can obtain improvement. If all training images adopt chain augmentation, the performances on clean and corrupted data can obtain a good balance.
In the meanwhile, if the random ratio is larger than 50\%, the model achieves approximate mIoU$_{c}$ scores.

\begin{table}[h]
\centering
\small
\caption{The effectiveness of single-scale and multi-scale testing with SCHP model on LIP-C dataset. }
\begin{tabular}{c | c  c }
\toprule
Method & Clean & mIoU$_{c}$ \\
\midrule
Single-scale & 58.62 & 43.39 \\
Single-scale+flipping & 59.00 & \textbf{43.74} \\
Multi-scale+flipping & \textbf{59.36} & 42.98 \\
\bottomrule

\end{tabular}

\label{tab:multiscale}
\end{table}

\begin{table}[h]
\centering
\small
\caption{The effectiveness of the adversarial attack on LIP dataset. }
\begin{tabular}{c | c | c c c }
\toprule
Method & Clean & PGD-10 & PGD-20 & PGD-40 \\
\midrule
SCHP \cite{li2020self} & 58.62 & 13.38  &	10.60 & 9.27 \\

SCHP-HeterAug & 57.95 & 18.97 & 14.56 & 11.76 \\
\bottomrule

\end{tabular}

\label{tab:pgd}
\end{table}

\textbf{The effectiveness of model-aware augmentation.} As shown in Table \ref{tab:deepaug}, if we adopt a deep neural network to generate perturbed images with random weights and control parameters, the robustness of the model can be enhanced. In particular, the percentage of 25\% achieves a good balance between clean and corrupted data.

\textbf{The effectiveness of composition type.} How do we combine these two augmentation operations with suitable combination types? We test two types, \textit{i.e.}, sequential augmentation and random sampling.
Sequential augmentation applies image-aware and model-aware augmentation operations sequentially. Random sampling means randomly choosing one operation from the set (original, chain augmentation, and perturbed deep network) implemented on each image according to pre-defined ratios. Specifically, we choose 50\% of images with chain augmentation, 25\% with the perturbed deep network, and 25\% are original images. As shown in Table \ref{tab:fuse}, the sequential augmentation achieves better performance than the random sampling one. Thus, we recommend sequential fusion to enrich the diversity of the training images.

\begin{table*}[t]
\footnotesize
\centering
\caption{Average clean and corrupted mIoU comparison results on Pascal-VOC2012-C dataset. * denotes three severity levels are considered.  }
\resizebox{\textwidth}{!}{
\begin{tabular}{ p{4.0cm} | c | c c c c | c c c c }
\toprule
\multirow{2}{*}{Method}  & \multirow{2}{*}{Clean} & \multicolumn{4}{c}{Blur}& \multicolumn{4}{|c}{Digital} \\
\cline{3-10}
~& ~& Defocus & Gaussian & Motion & Glass & Brightness & Contrast & Saturate & JPEG \\
\hline

Deeplabv3plus(Res50)\cite{kamann2021benchmarking}              & 69.60  & 43.50 & 45.50 & 38.70 & 31.10 & 63.50 & 50.30 & 63.80 & 58.20     \\
Deeplabv3plus(Res101-d16)\cite{kamann2021benchmarking}             & 70.30  & 45.60 & 46.60 & \textbf{45.80} & 33.20 & 64.50 & 50.60 & 65.30 & 59.70     \\
Deeplabv3plus(Res101-d8)             & 77.82  & 48.27 & 49.86 & 42.64 & 34.24 & 71.60 & 47.22 & 70.10 & 59.34     \\
PSPNet \cite{zhao2017pyramid} & 77.80 & 49.28 & 51.42 & 43.41 & 34.69  & 70.51&	46.67	&68.60	&58.54 \\
\hline
Deeplabv3plus-HeterAug & 77.21 & \textbf{52.69} & 54.21 & 42.52 & 35.12 & 74.72 & 62.33 & 74.24 & \textbf{67.09} \\

PSPNet-HeterAug  & \textbf{78.31} & 51.94 & \textbf{54.23} & 40.70 & \textbf{36.73}  & \textbf{75.63} & \textbf{62.78} & \textbf{75.17} & 65.63 \\
\hline
\end{tabular}
}\footnotesize

\resizebox{\textwidth}{!}{
\begin{tabular}{ p{3.75cm} | c  c c c |c  c c c | c }
\toprule
\multirow{2}{*}{Method}  & \multicolumn{4}{c}{Noise*}& \multicolumn{4}{|c|}{Weather} & \multirow{2}{*}{mIoU$_{c}$} \\
\cline{2-9}
~ &  Gaussian & Impulse & Shot & Speckle & Fog & Frost & Snow & Spatter & ~ \\
\hline

Deeplabv3plus(Res50)\cite{kamann2021benchmarking}               & 43.20 & 40.70 & 44.20 & 50.90 & 56.90 & 39.80 & 31.30 & 47.00 & 46.79 \\
Deeplabv3plus(Res101-d16)\cite{kamann2021benchmarking}          & 49.40 & 48.30 & 50.10 & 55.40 & 57.60 & 41.20 & 31.40 & 50.40 & 49.69 \\
Deeplabv3plus(Res101-d8)           & 51.99 & 52.46 & 52.35 & 58.83 & 63.70 & 47.96 & 39.63 & 58.95 & 53.07 \\
PSPNet \cite{zhao2017pyramid} & 43.92 & 43.90 & 45.43 & 55.12 & 62.28 & 43.85 & 36.50 & 56.27 & 50.65 \\
\hline
Deeplabv3plus-HeterAug & 64.68 & 61.88 & 65.95 & 69.03 & 69.97 & 53.54 & 41.74 & 62.81 & 59.53 \\

PSPNet-HeterAug  & \textbf{65.40} & \textbf{62.96} & \textbf{66.84}  & \textbf{70.65} & \textbf{70.51} & \textbf{54.99} & \textbf{43.72} & \textbf{63.13} & \textbf{60.06}  \\
\hline
\end{tabular}
}

\label{tab:voc}
\end{table*}

\textbf{The effectiveness of mix fusion of image-aware augmentation.} Though the chain augmentation has obtained good performance on the corrupted data, we find that the mix type for image-aware augmentation can further improve the robustness performance. The robust model should not only keep the ability of the clean data but also owns high generality on the unforeseen image common corruptions. Thus, we propose a mixed fusion that samples weights from Beta distribution combining the original image and the chain augmented image to generate a new image. The experimental results in Table \ref{tab:mix} show the mixed type (ImageAug) can obtain better performance ($43.81\%$) than the best chain augmentation case (43.25\% in Table \ref{tab:chain}). The advantage of the mixed fusion lies in that it not only obtains a better mIoU score but also effectively avoids hyper-parameter (\textit{i.e.}, random ratio) influence.

Finally, the image-aware and the model-aware augmentations are fused in a sequential manner, the results are shown in Table \ref{tab:mix}. We can observe that the sequential fusion (HeterAug) obtains 1.22\% and 1.33\% gains over image-aware and model-aware augmentations, respectively. The heterogeneous augmentation method on the clean data  achieves better performance than the baseline method CE2P, and the performance is approximate to image-aware and model-aware augmentations.
In addition, we also try another augmentation method named grid-mask \cite{chen2020gridmask}, which randomly removes the image region to enhance the learning difficulty. And the grid mask mechanism has been used in some computer vision tasks to improve performance. While we adopt the grid mask augmentation, we find it is also a useful clue for improving the robustness of the model, as shown in Table \ref{tab:mix}. The grid mask augmentation (+Grid Mask) can obtain a 2.31\% gain over the CE2P baseline. The performance gain of grid mask augmentation is lower than that of image-aware and model-aware augmentations. Thus, we recommend the heterogeneous augmentation strategy to improve the robustness of the human parsing models.
The heterogeneous augmentation mechanism obtains a 6.62\% robust gain over the CE2P baseline.
Besides, the proposed HeterAug does not bring extra computational time. We test the inference time of the SCHP and SCHP\_HeterAug models on the validation set. The running time are 17.08 and 17.12 FPS respectively. Therefore, the proposed HeterAug mechanism exhibits excellent performance and does not incur excessive computational overhead.

\subsection{The Influence of Multi-scale Testing}

In general, multi-scale testing is a good choice to help the model to obtain a better performance. We also verify it in corrupted conditions. We evaluate three testing types with SCHP\cite{li2020self} on the LIP-C dataset,
\textit{i.e.}, Single-scale, Single-scale+horizontal flipping, and Multi-scale+horizontal flipping, the results are shown in Table \ref{tab:multiscale}. The multi-scale testing contains 0.5, 0.75, 1, 1.25, and 1.5. From Table \ref{tab:multiscale}, the multi-scale testing does not bring performance gain, which is a little worse than the single-scale testing. The main weakness is related to the low quality of the corrupted images. Thus, we recommend the Single-scale+horizontal flipping for corrupted images in the inference stage.

\subsection{The Influence of Adversarial Attack}

We also assess the effectiveness of our proposed method on the adversarial attack. Specifically, we choose the PGD attack \cite{madry2017towards} with different iterations for comparison, as shown in Table \ref{tab:pgd}. We can find that the performances of both the SCHP and SCHP-HeterAug methods drop. Benefiting from the heterogeneous augmentation strategy, the SCHP-HeterAug method obtains better performance than the SCHP, which shows the effectiveness of our proposed HeterAug method. However, the performance drop is also large, thus we need to further improve the robustness of the human parsers by taking the adversarial attack corruption into account.

\begin{figure*}[t]
  \centering
  \subfigure[Defocus blur.]{
    \label{fig:defocusblur}
    \includegraphics[width=0.23\linewidth]{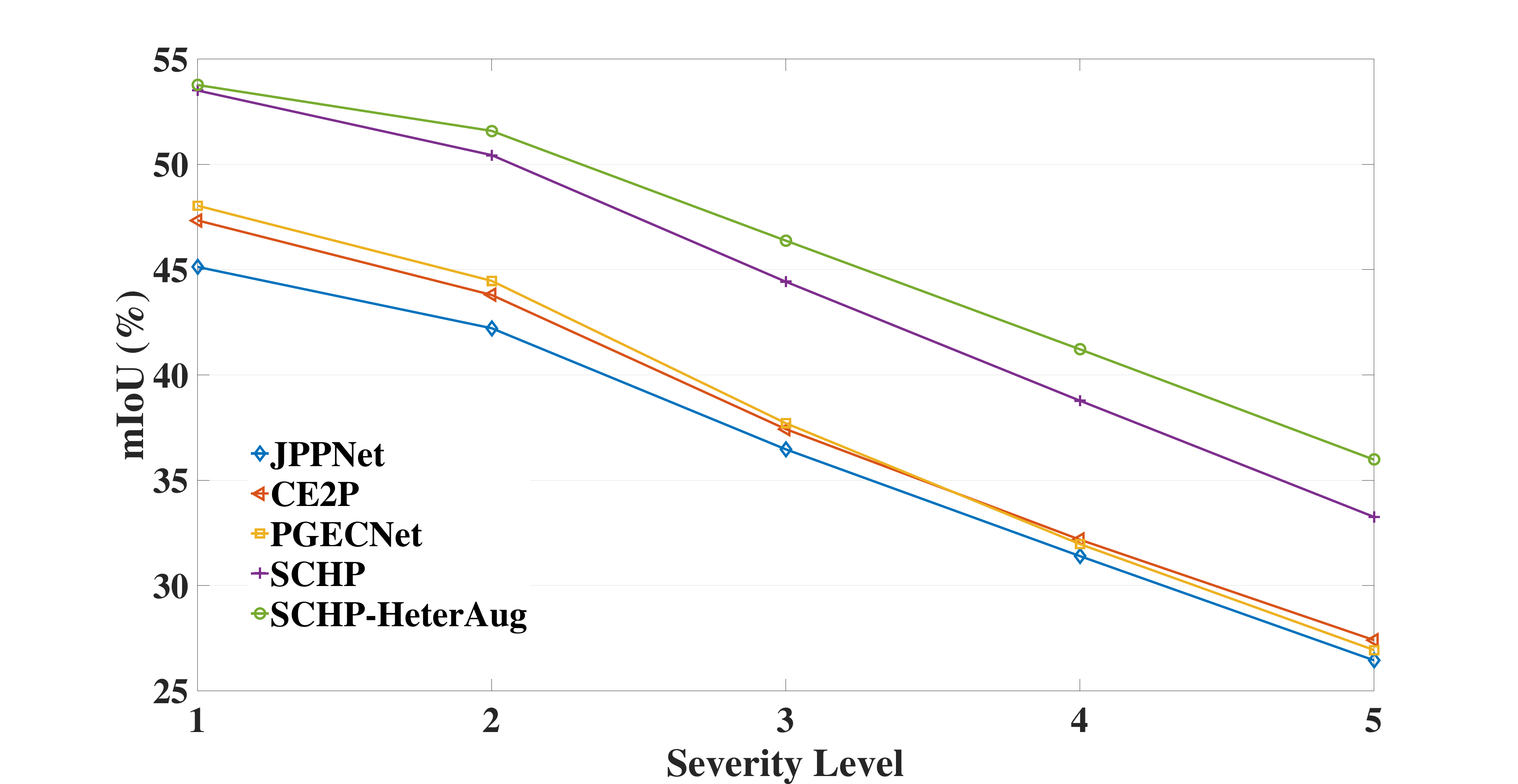}}
    \hfill
    \subfigure[Gaussian blur.]{
    \label{fig:gaussianblur}
    \includegraphics[width=0.23\linewidth]{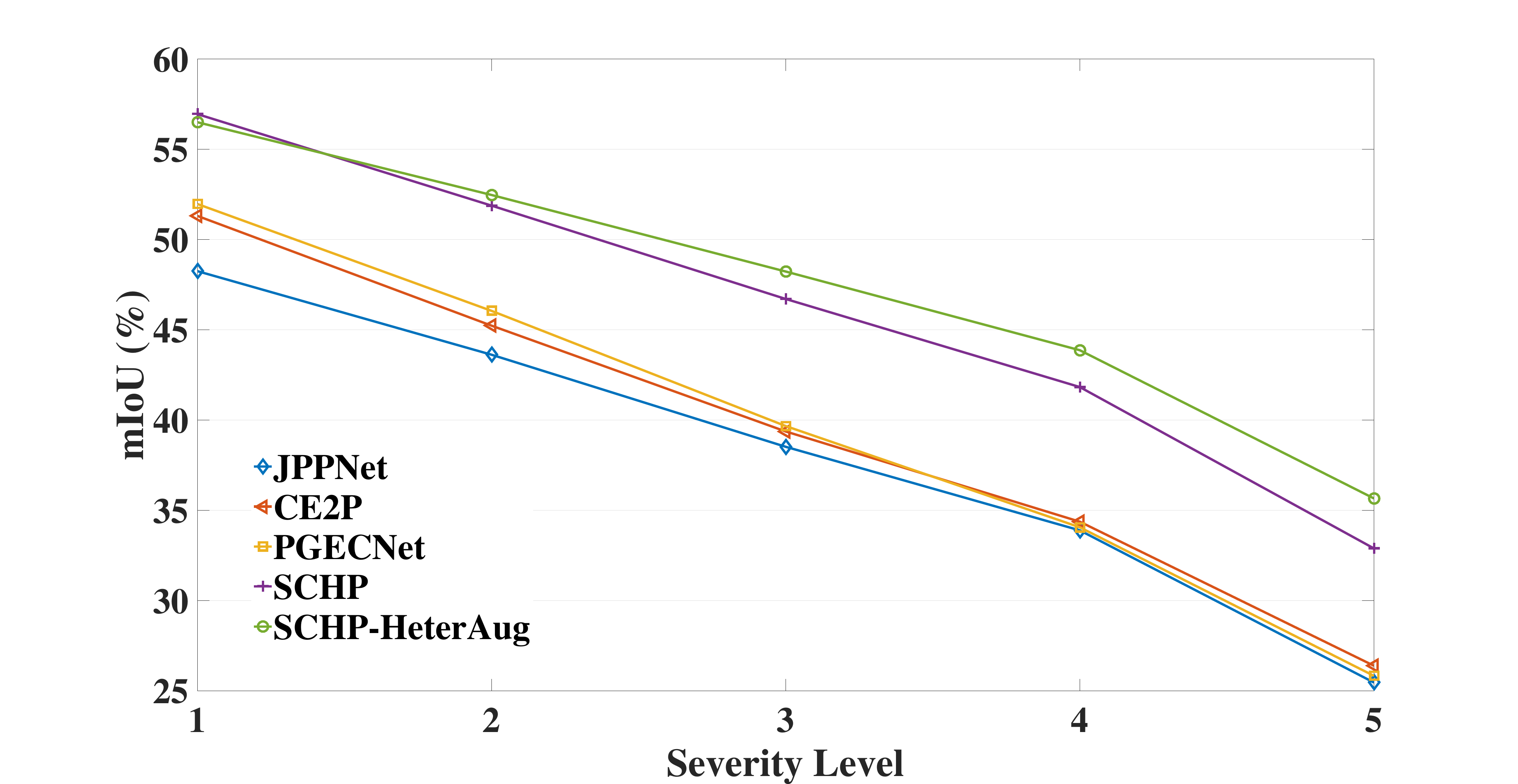}}
    \hfill
    \subfigure[Motion blur.]{
    \label{fig:motionblur}
    \includegraphics[width=0.23\linewidth]{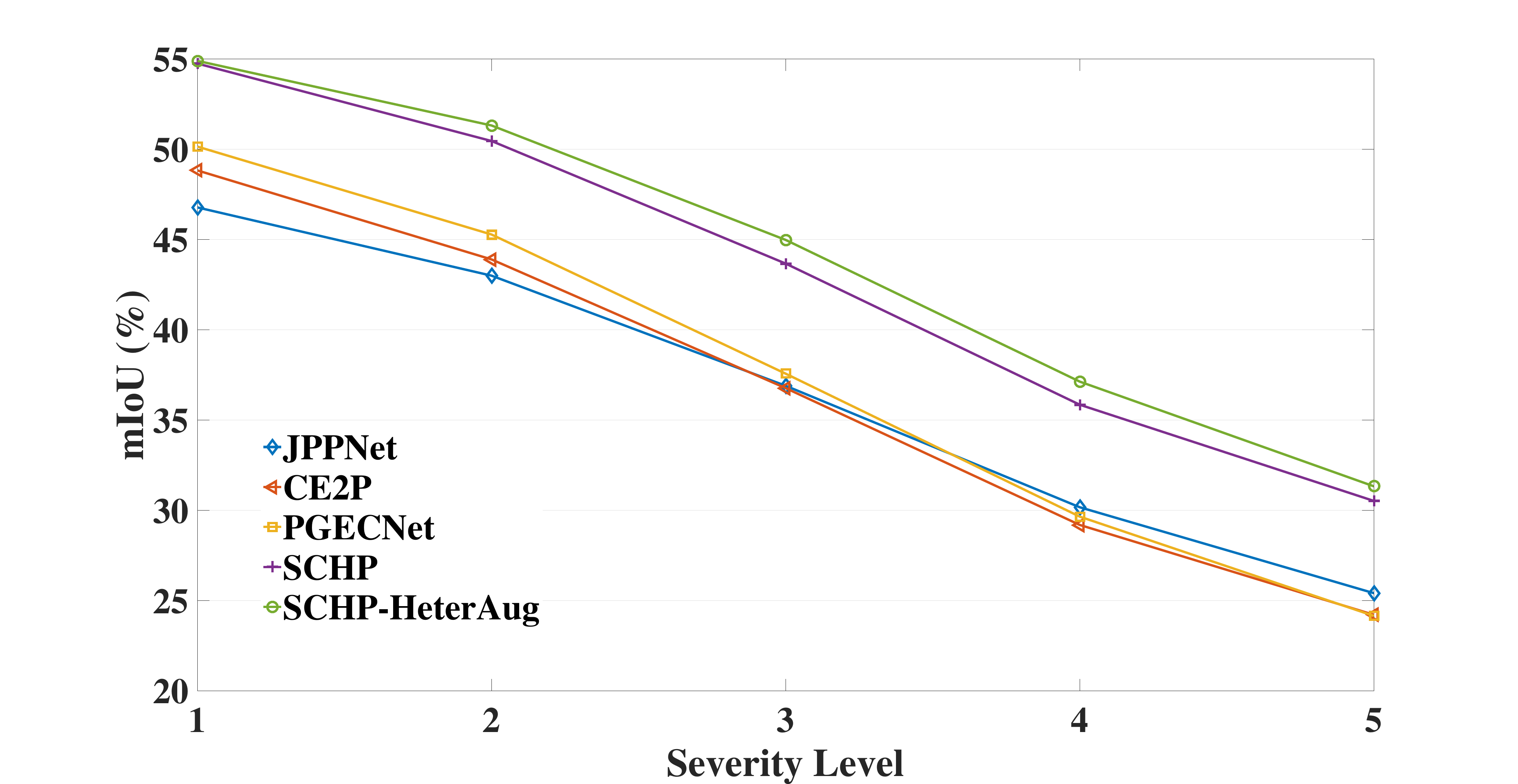}}
    \hfill
    \subfigure[Glass blur.]{
    \label{fig:glassblur}
    \includegraphics[width=0.23\linewidth]{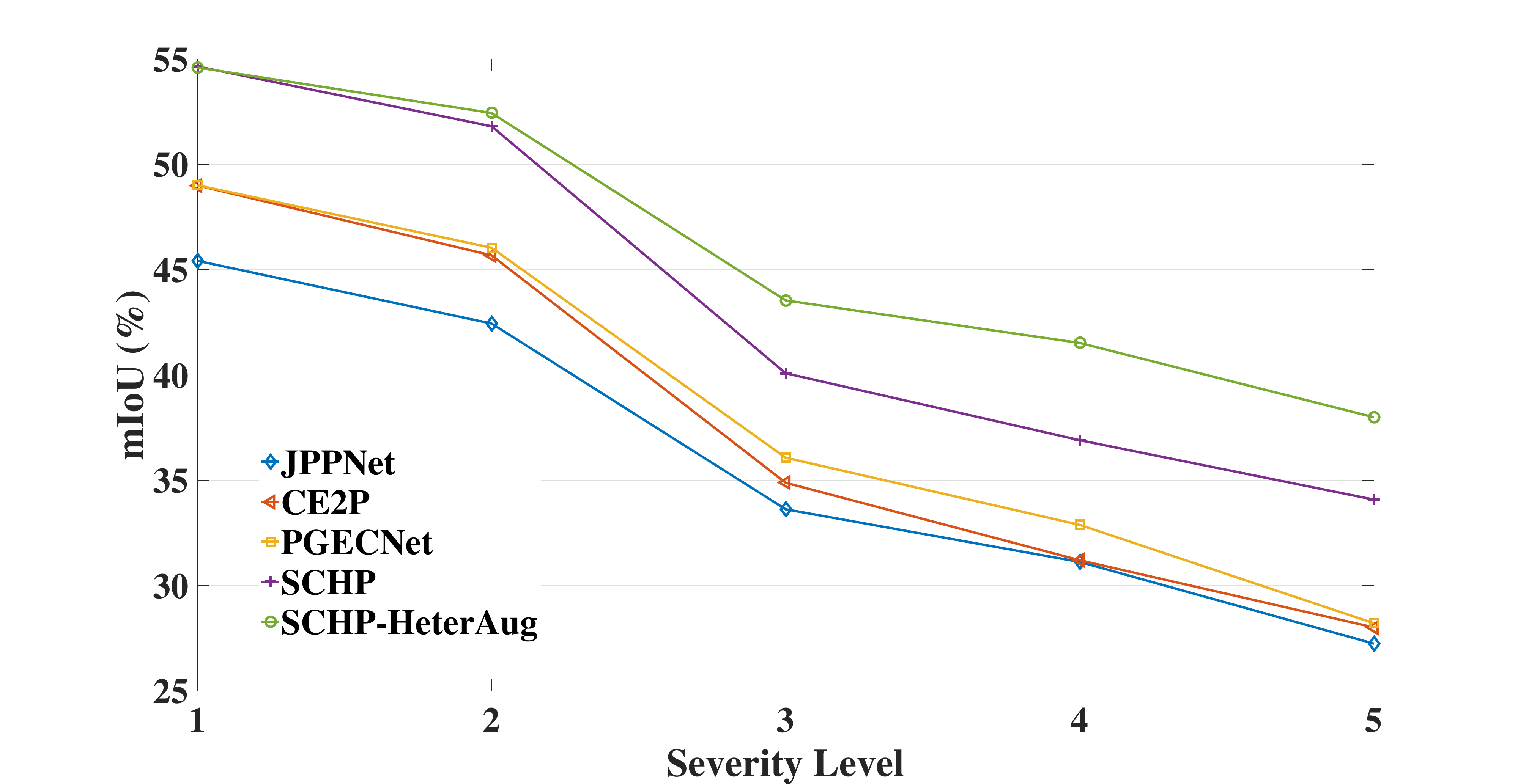}}

      \subfigure[Brightness.]{
    \label{fig:brightness}
    \includegraphics[width=0.23\linewidth]{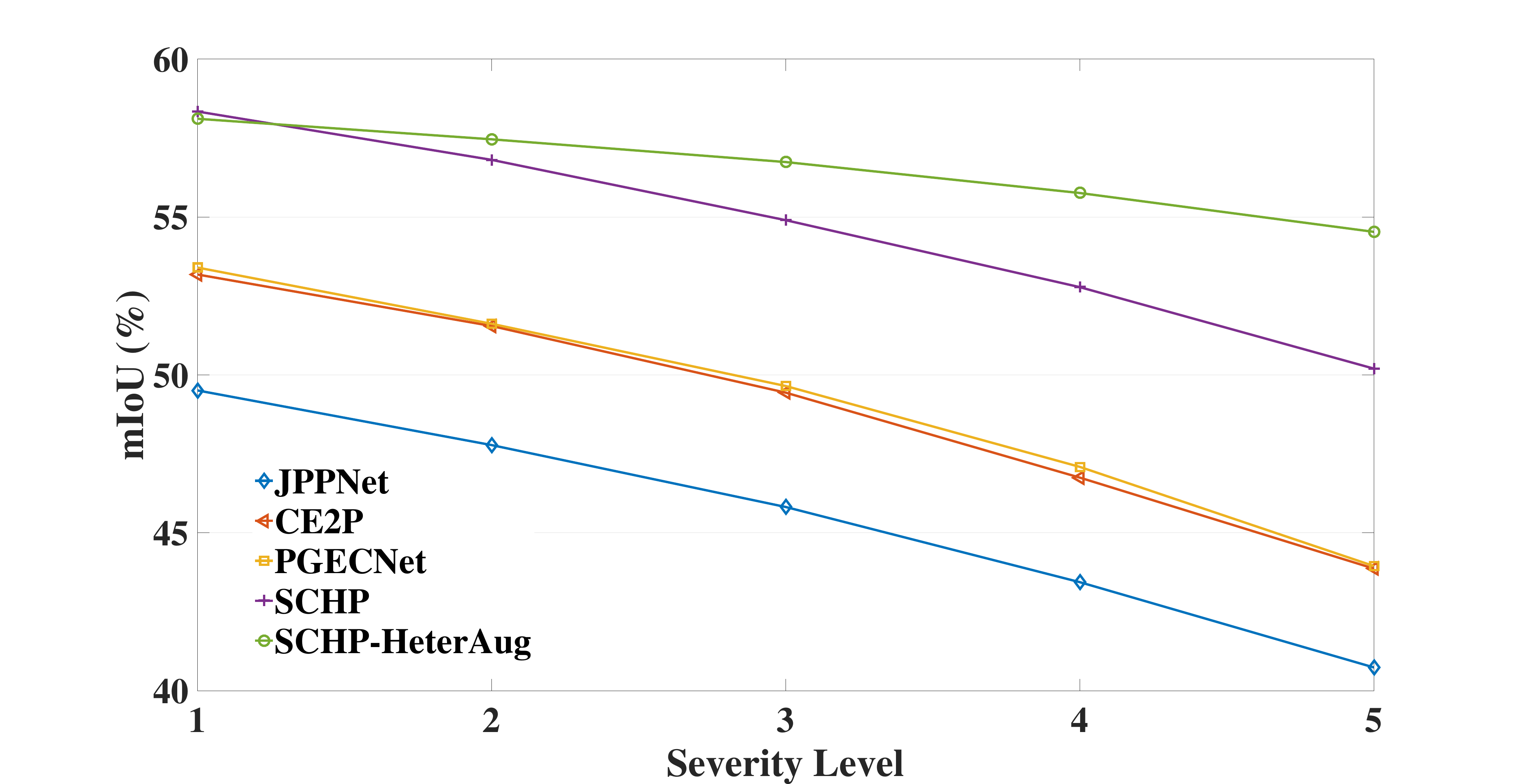}}
    \hfill
    \subfigure[Contrast.]{
    \label{fig:contrast}
    \includegraphics[width=0.23\linewidth]{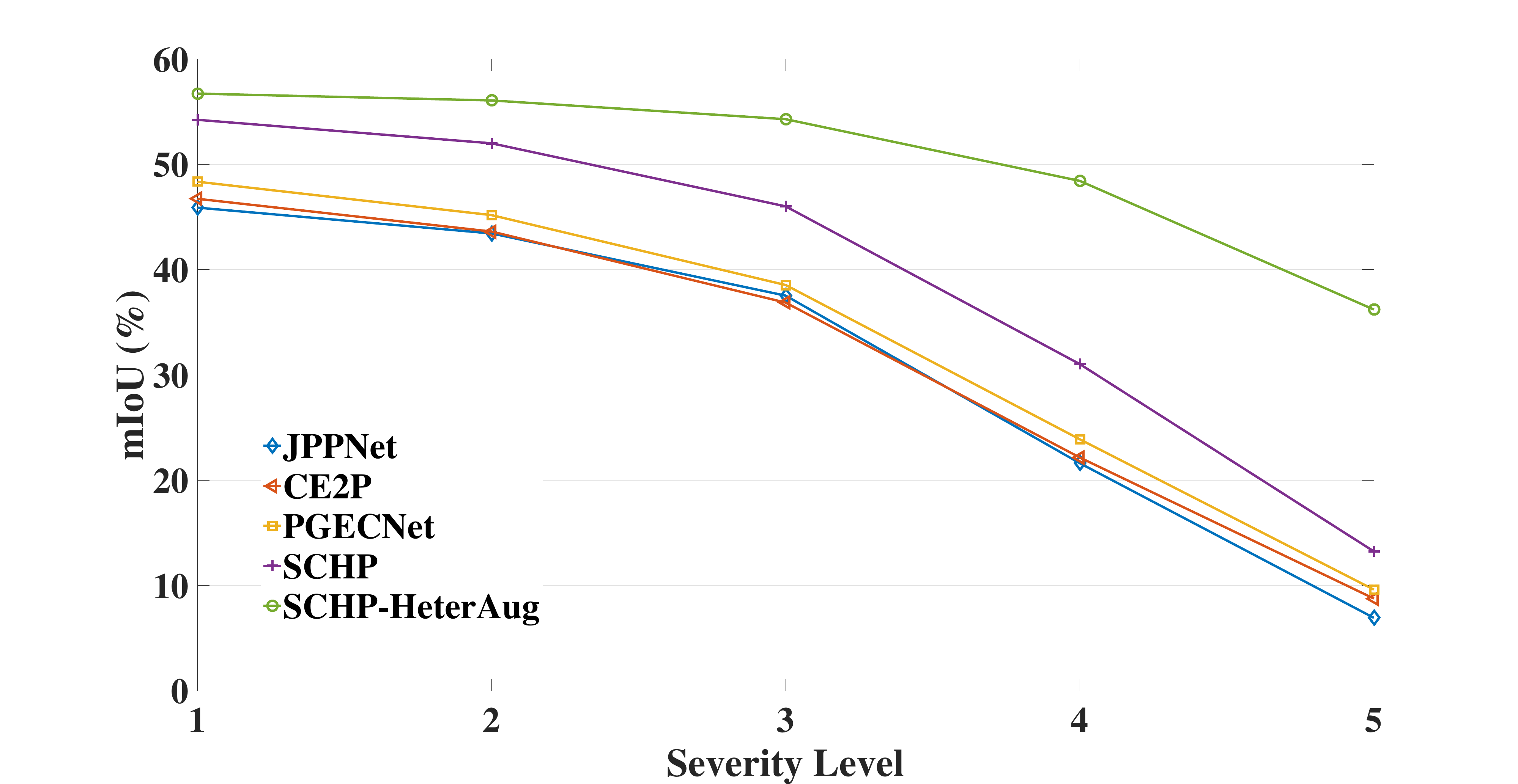}}
\hfill
    \subfigure[Saturate.]{
    \label{fig:saturate}
    \includegraphics[width=0.23\linewidth]{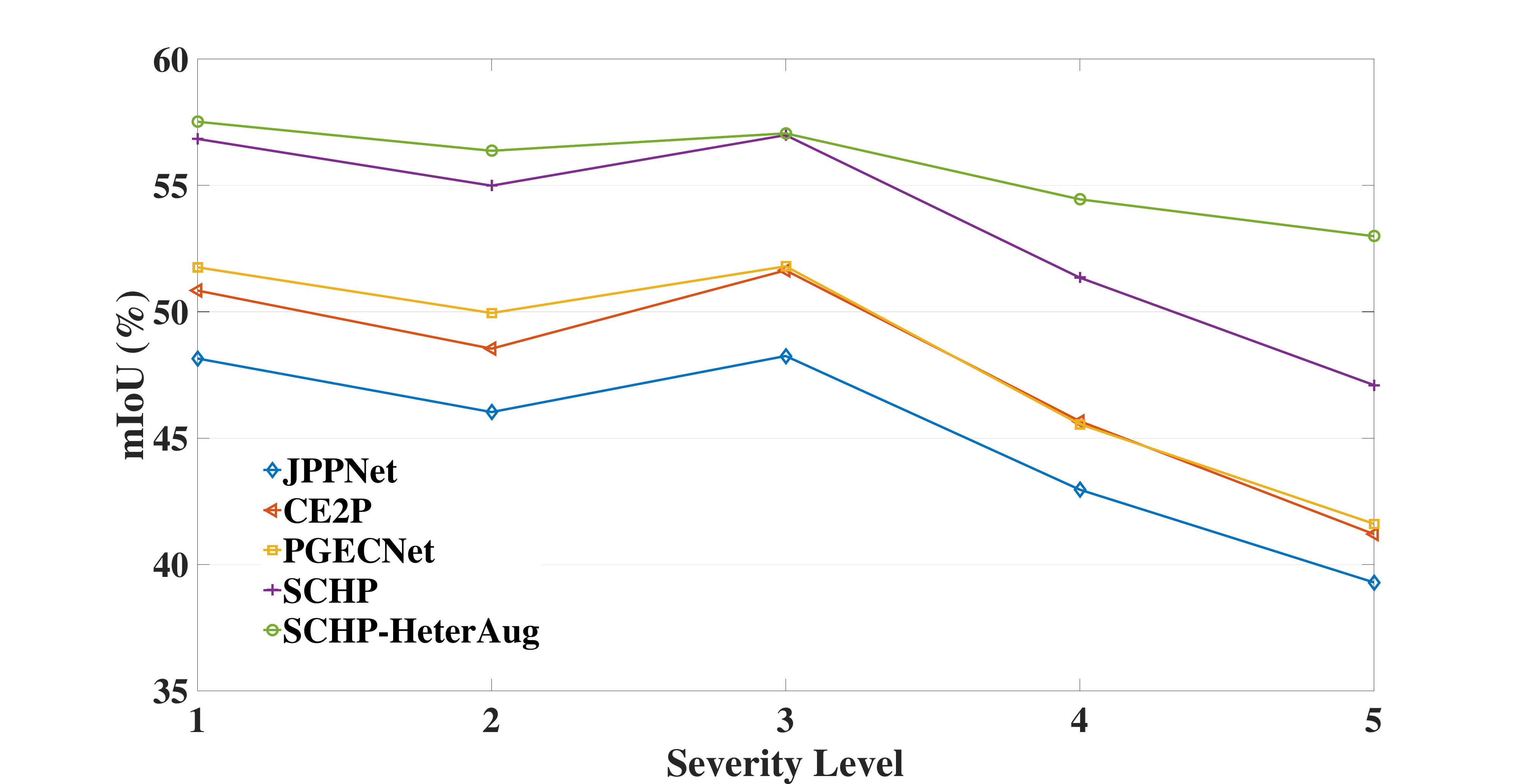}}
    \hfill
    \subfigure[JPEG compression.]{
    \label{fig:jpegcompression}
    \includegraphics[width=0.23\linewidth]{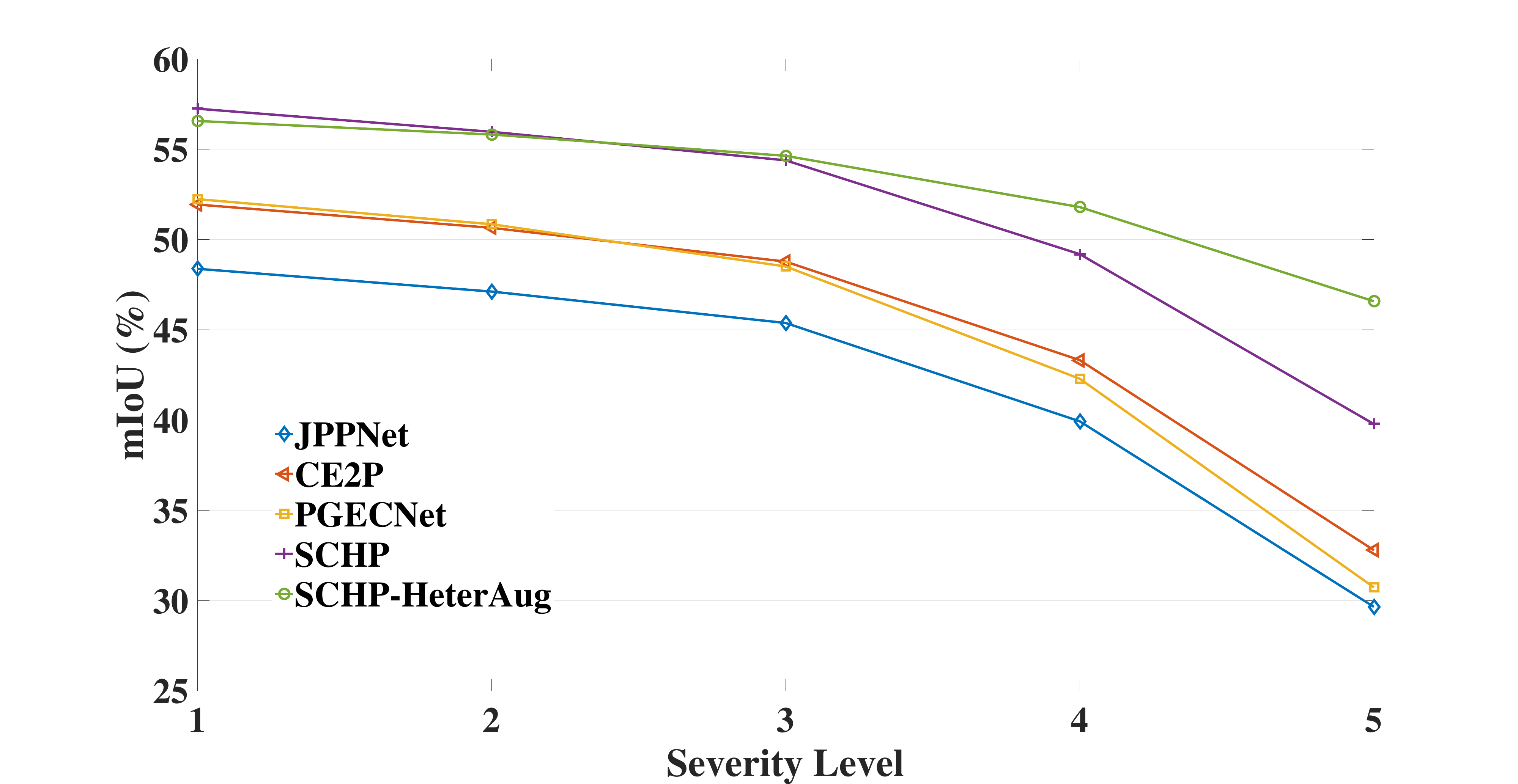}}

  \subfigure[Gaussian noise.]{
    \label{fig:gaussiannoise}
    \includegraphics[width=0.23\linewidth]{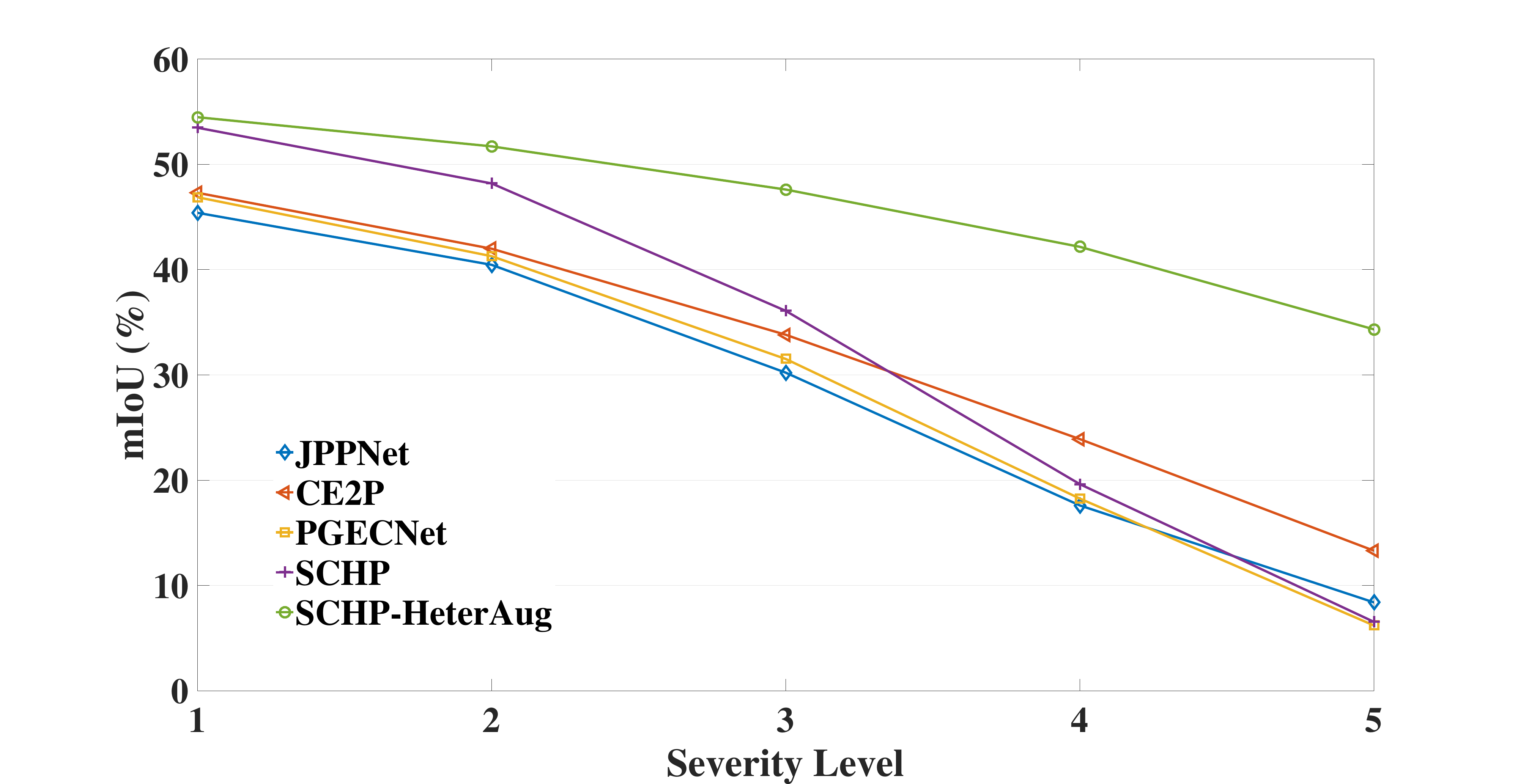}}
    \hfill
    \subfigure[Impulse noise.]{
    \label{fig:impulsenoise}
    \includegraphics[width=0.23\linewidth]{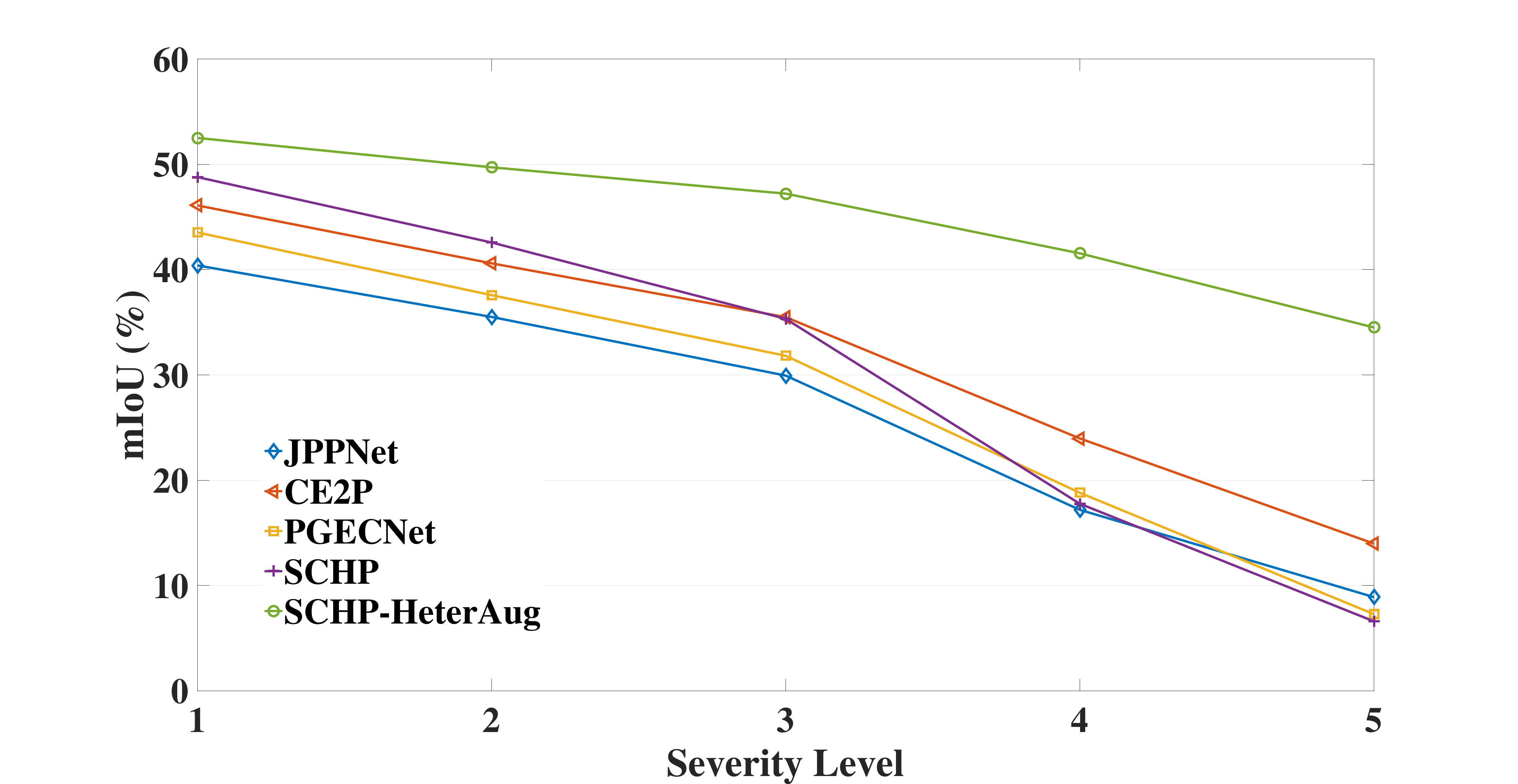}}
    \subfigure[Shot noise.]{
    \label{fig:shotnoise}
    \includegraphics[width=0.23\linewidth]{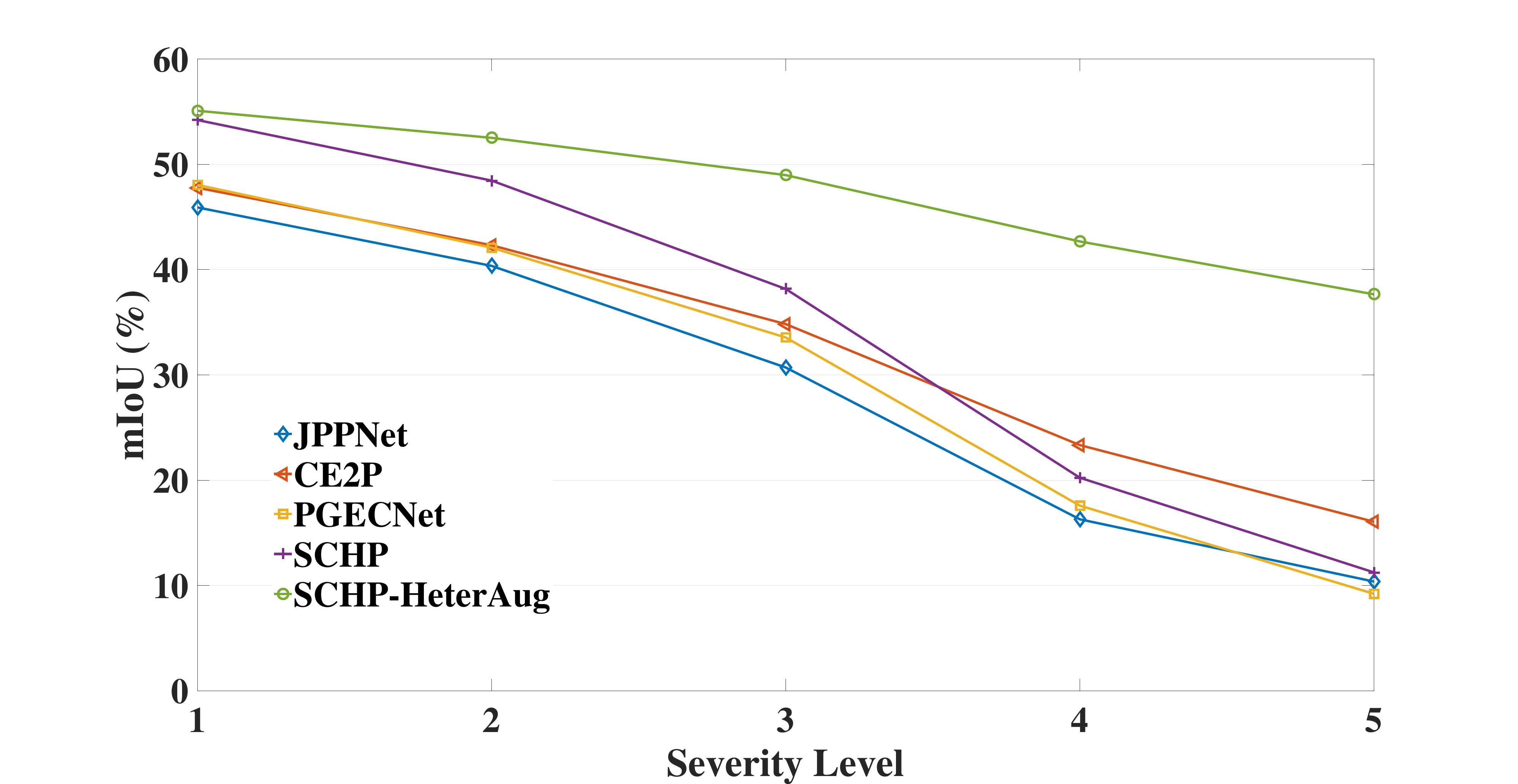}}
    \hfill
    \subfigure[Speckle noise.]{
    \label{fig:specklenoise}
    \includegraphics[width=0.23\linewidth]{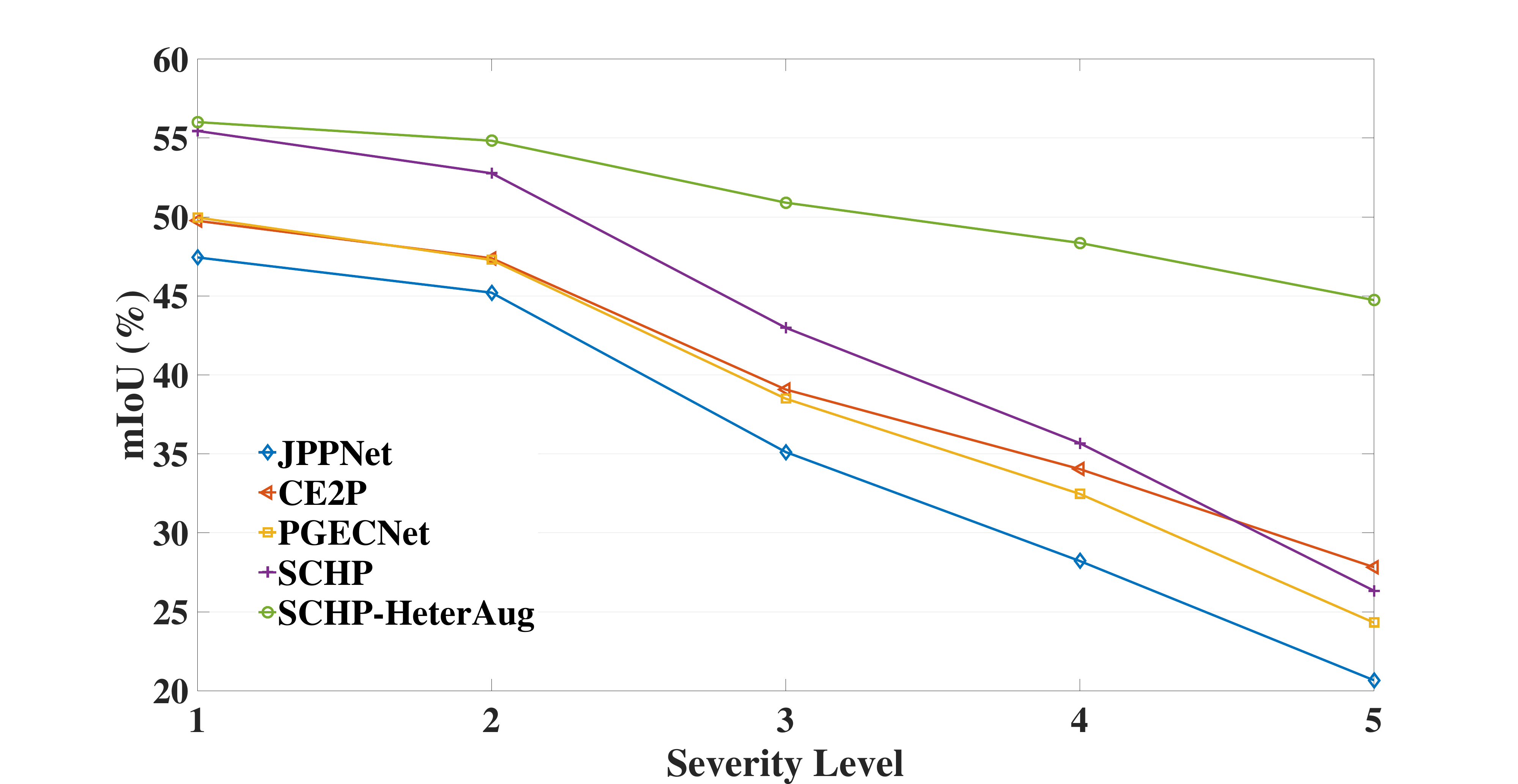}}

  \subfigure[Fog.]{
    \label{fig:fog}
    \includegraphics[width=0.23\linewidth]{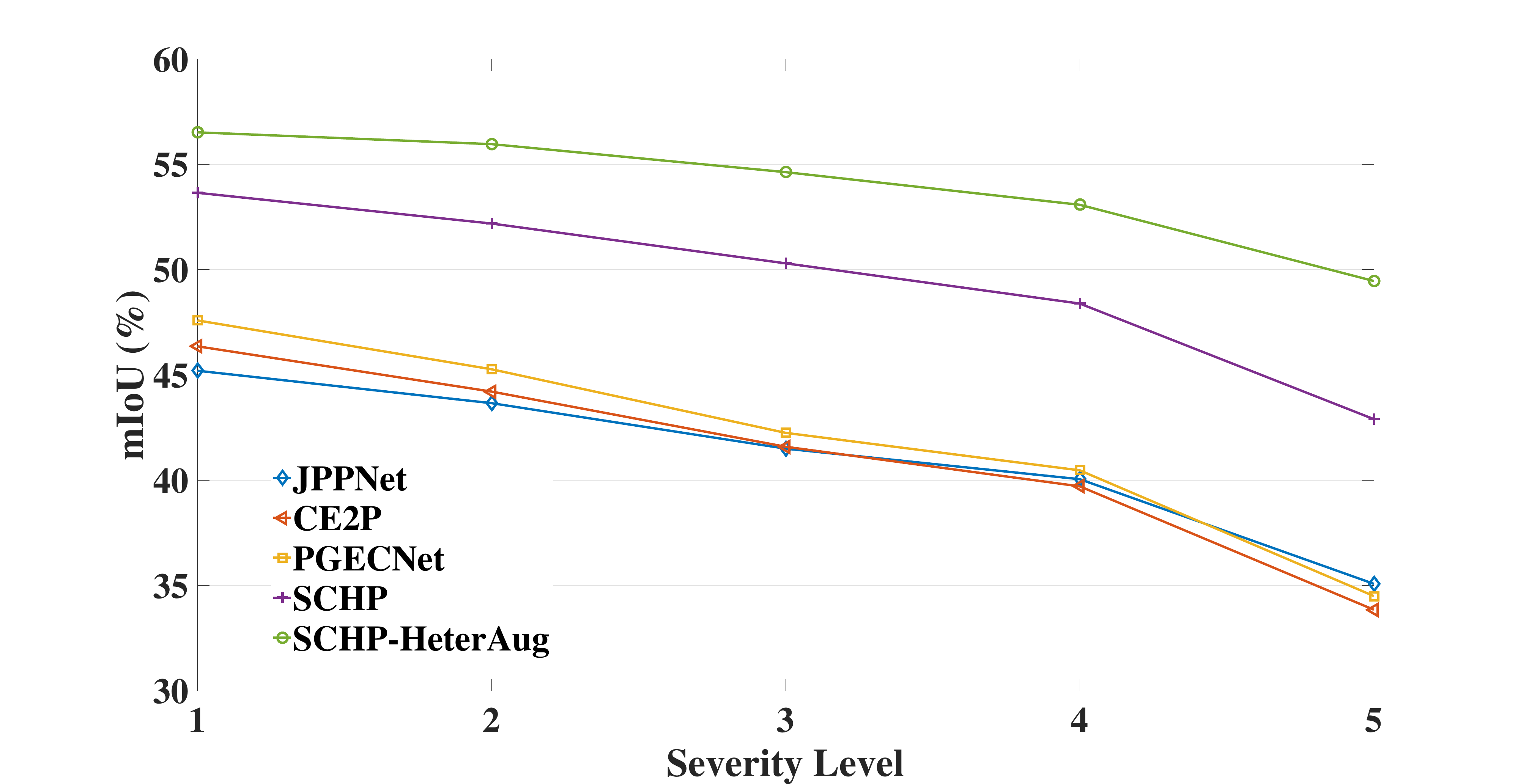}}
    \hfill
    \subfigure[Frost.]{
    \label{fig:frost}
    \includegraphics[width=0.23\linewidth]{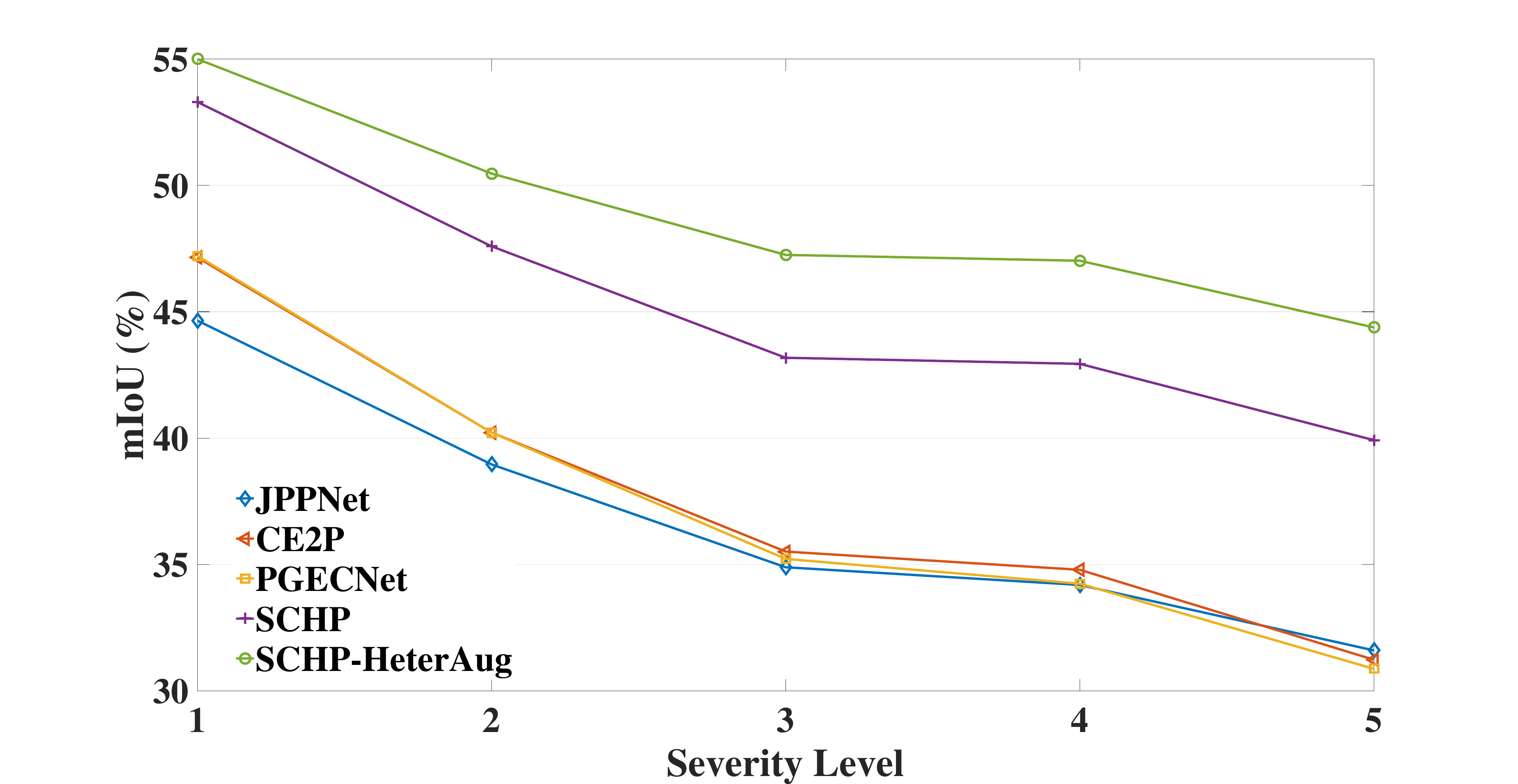}}
    \subfigure[Snow.]{
    \label{fig:snow}
    \includegraphics[width=0.23\linewidth]{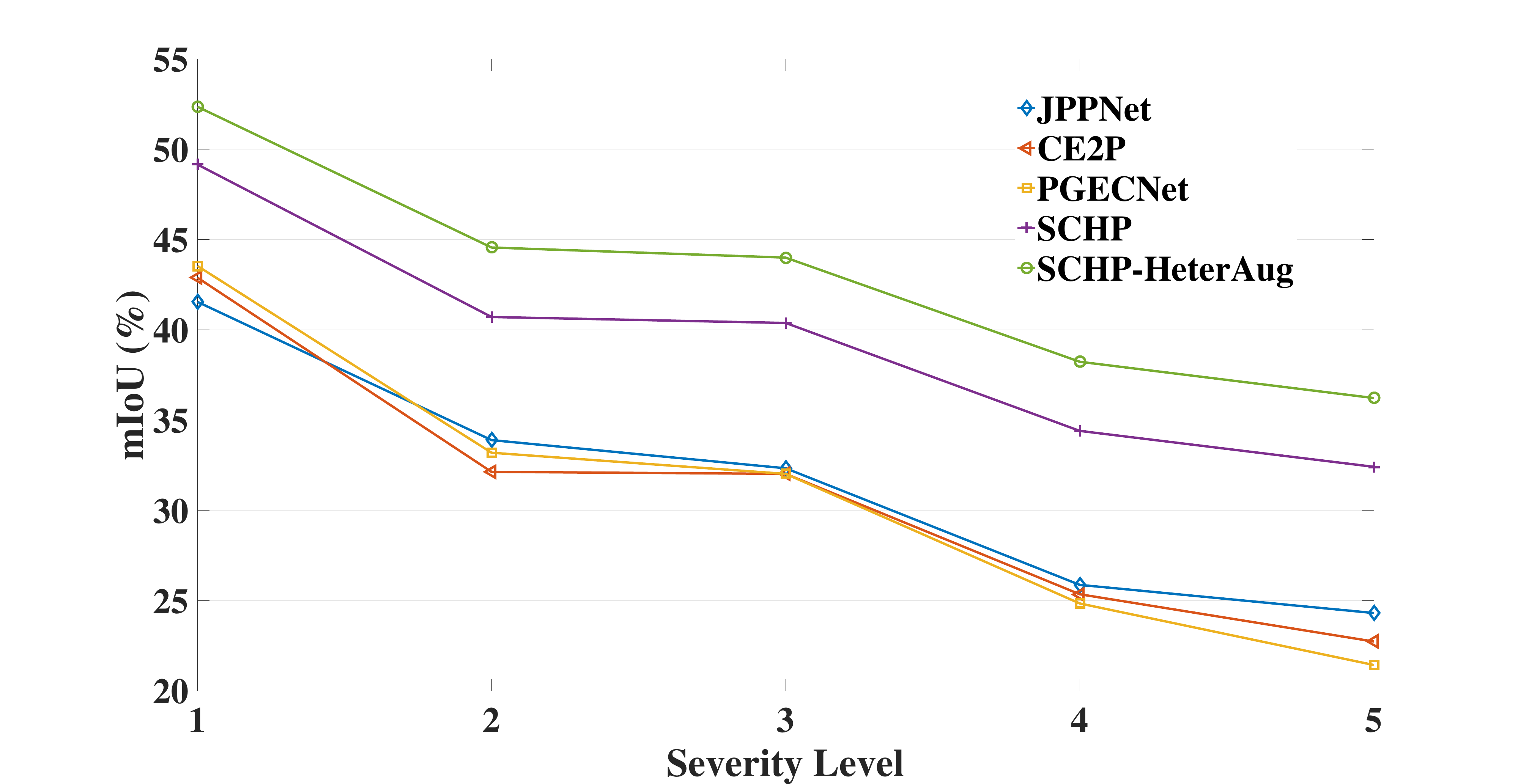}}
    \hfill
    \subfigure[Spatter.]{
    \label{fig:spatter}
    \includegraphics[width=0.23\linewidth]{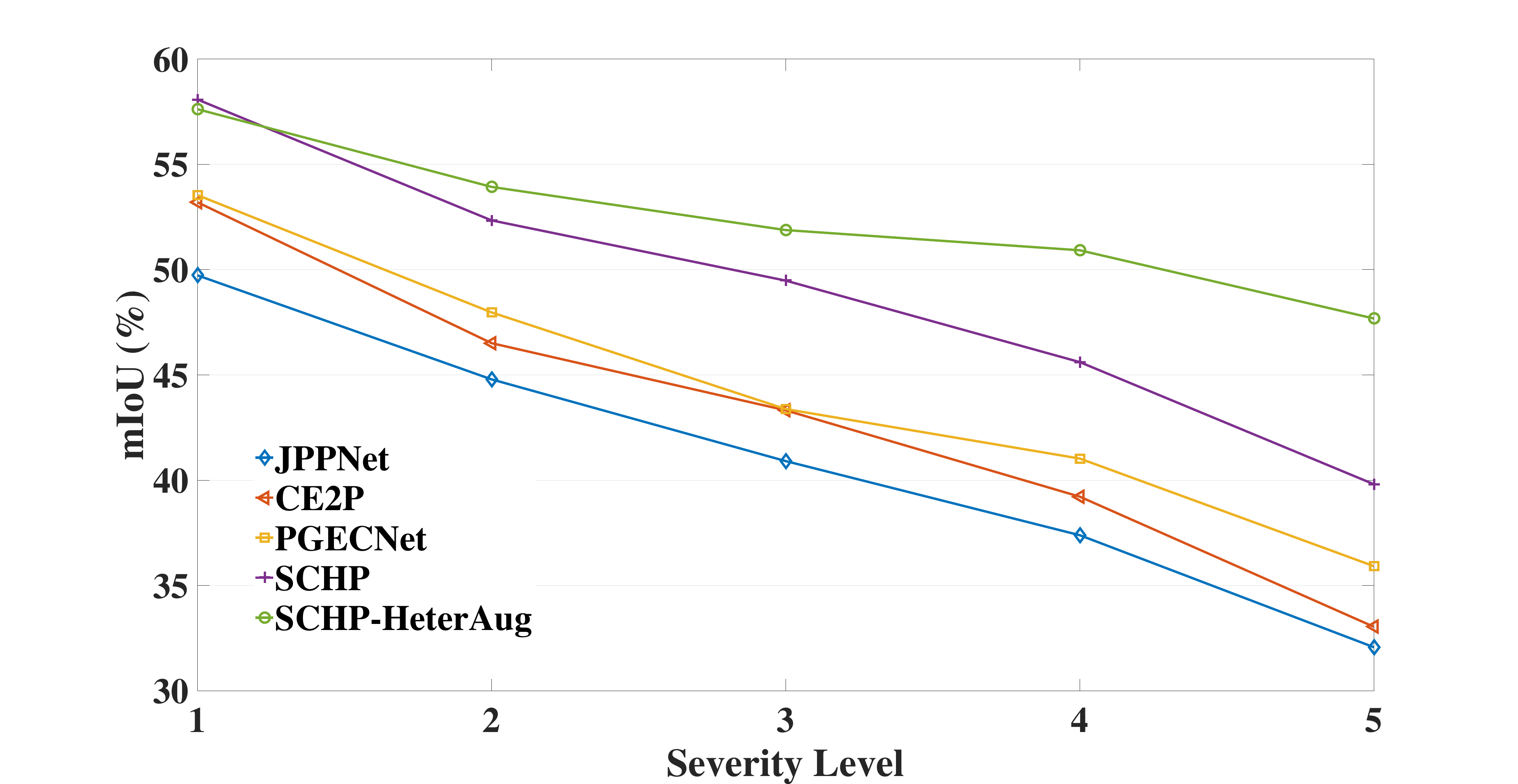}}

  \caption{Model performance (mIoU) for different severity levels (1-5) with respect to the image corruption categories, \textit{i.e.}, blur (defocus, gaussian, motion, and glass), digital (brightness, contrast, saturate, and jpeg compression), noise (gaussian, impulse, shot, and speckle) and weather (fog, frost, snow, and spatter) on the LIP-C dataset.}
  \label{fig:level}
\end{figure*}

\subsection{Generalization on Pascal VOC 2012 Dataset}

To verify the effectiveness of the general semantic segmentation task, we apply the heterogeneous augmentation strategy to the Pascal VOC 2012 dataset \cite{everingham2010pascal}. The Pascal VOC 2012 dataset contains 21 classes, we adopt the augmented train set to train and report the mIoU scores on the validation set.
We choose the Deeplabv3plus \cite{chen2018encoder} and PSPNet \cite{zhao2017pyramid} to verify the effectiveness of the semantic segmentation model, and we adopt the open-source library \textbf{MMSegmentation}\footnote{https://github.com/open-mmlab/mmsegmentation} to implement it. The backbone is the ResNet-101, and the output stride is 8. The crop size is set as $512 \times 512$. The total iteration is set as $40,000$.

The comparison results on the Pascal VOC 2012 dataset are shown in Table \ref{tab:voc}.
For the sake of a fair comparison, we adopt the same experimental setup as \cite{kamann2021benchmarking}. The only distinction lies in the severity level of noise corruption, which is now set at 3. We have incorporated the heterogeneous augmentation mechanism into the deeplabv3plus model with an output stride of 8. This modified version is denoted as Deeplabv3plus(Res101-d8). In most common corruptions, the method Deeplabv3plus(Res101-d8) achieves better mIoU scores than Deeplabv3plus(Res101-d16).
If we introduce the heterogeneous augmentation mechanism into the Deeplabv3plus and PSPNet models, the HeterAug obtains large gains (5.96\% and 9.41\%) upon the deeplabv3plus(Res101-d8) and PSPNet on the VOC dataset. Thus, the good performance further shows the generality of our proposed heterogeneous augmentation method. The main advantage lies in that the training samples are enriched through image-aware and model-aware augmentation mechanisms.

\subsection{Qualitative Results}

\textbf{The effectiveness of different severity levels.}
In this section, we discuss the influences of different severity levels implemented on image corruption.
Specifically, five methods are employed to compare, \textit{i.e.}, JPPNet \cite{liang2018look}, CE2P \cite{ruan2019devil}, PGECNet \cite{zhang2021human}, SCHP \cite{li2020self} and SCHP-HeterAug (Ours). The results of 16 image corruptions with 5 severity levels testing on the LIP-C dataset are shown in Figure \ref{fig:level}.
By observing the results, we find that different severity levels also influence the performance of the human parsing model. In general, the larger severity level will bring worse performance. For example, in the contrast corruption in Figure \ref{fig:contrast},
when the severity level is larger than 3, the performance of comparison methods drops faster than prior levels. This indicates that the current state-of-the-art human parsing models are highly sensitive to the contrast corruption. Benefiting from our proposed heterogeneous augmentation strategy, the performances of levels 4 and 5 drop lower than other methods. A similar observation can also be found on the shot noise corruption in Figure \ref{fig:shotnoise}. These results show that the heterogeneous augmentation helps the model deal with these different images' common corruptions well.

\begin{figure}[h]
  \centering
  \includegraphics[width=0.8\linewidth]{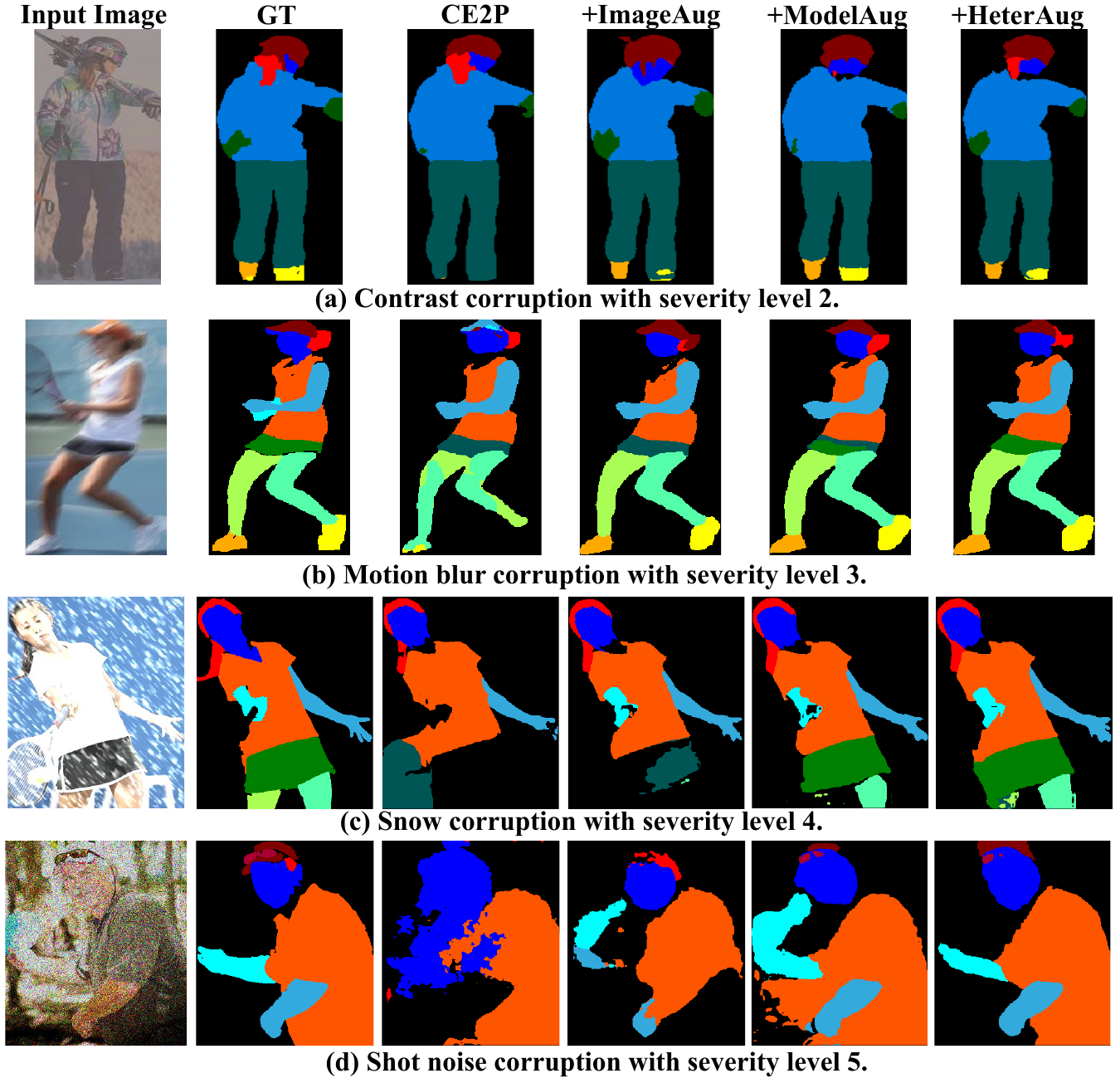}
  \caption{ The qualitative effectiveness of the image-aware and model-aware augmentations on corrupted LIP-C dataset.
  }
  \label{fig:ablation}
\end{figure}

\textbf{Qualitative comparisons of the image-aware and model-aware augmentation mechanisms.}
We choose CE2P as the baseline and implement the image-aware, model-aware and heterogeneous augmentations, perspectively, as shown in Figure \ref{fig:ablation}. In Figure \ref{fig:ablation} (c) and (d), both adding either image-aware or model-aware augmentations can improve the parsing abilities compared with the CE2P method, and the HeterAug mechanism further obtains better results. 
CE2P struggles to predict certain small classes, like the left shoe and right shoe, as they share similarities with the background. 
Nevertheless, leveraging augmentation methods proves beneficial in achieving improved prediction results for these classes, as shown in Figure \ref{fig:ablation} (a) and (b).

\begin{figure*}
	\begin{center}
		\begin{tabular}{@{}c@{}}
			\includegraphics[width = 0.85\textwidth]{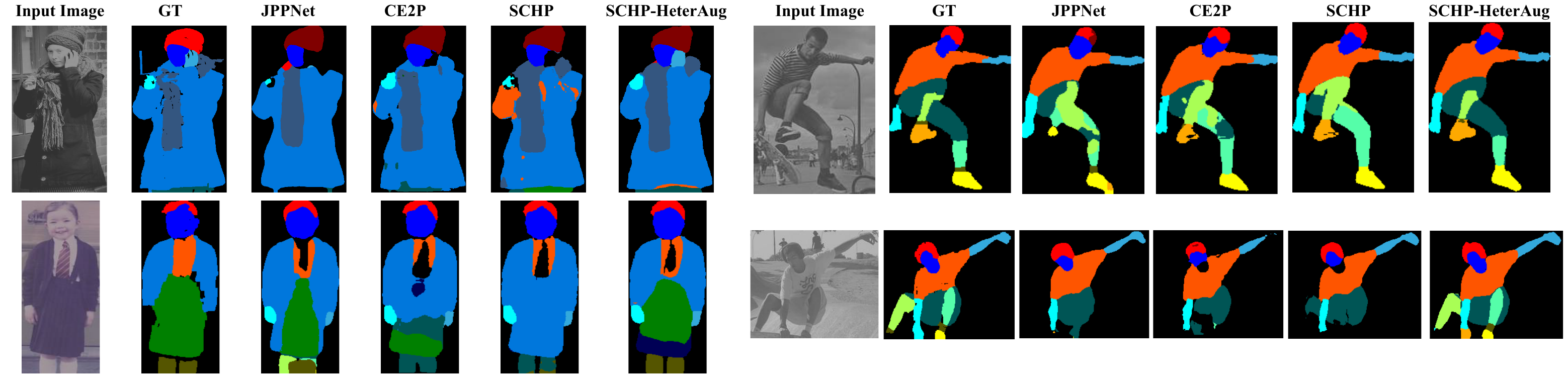} \\
			(a) Contrast corruption with severity level 1. \\
			\includegraphics[width = 0.85\textwidth]{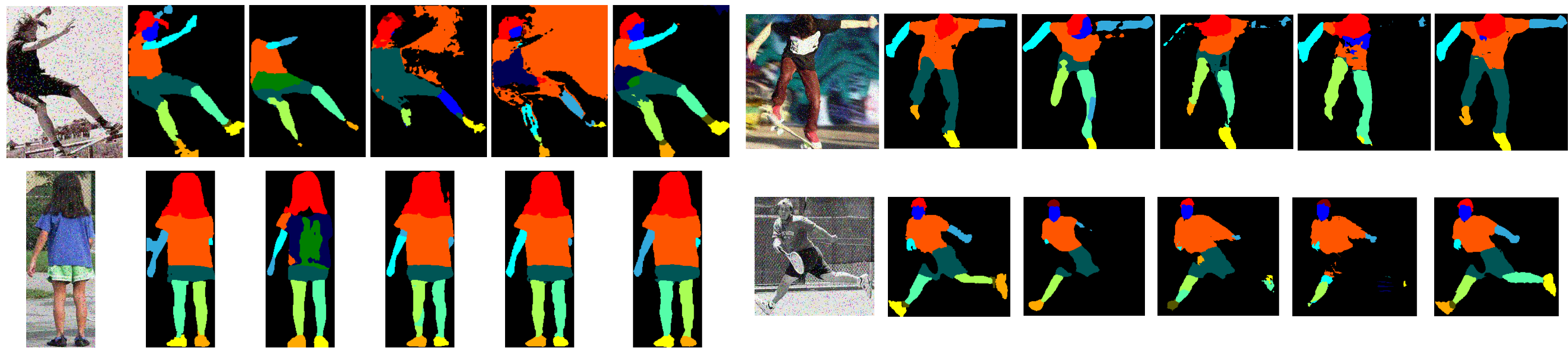} \\
			(b) Impulse noise corruption with severity level 2. \\
			\includegraphics[width = 0.85\textwidth]{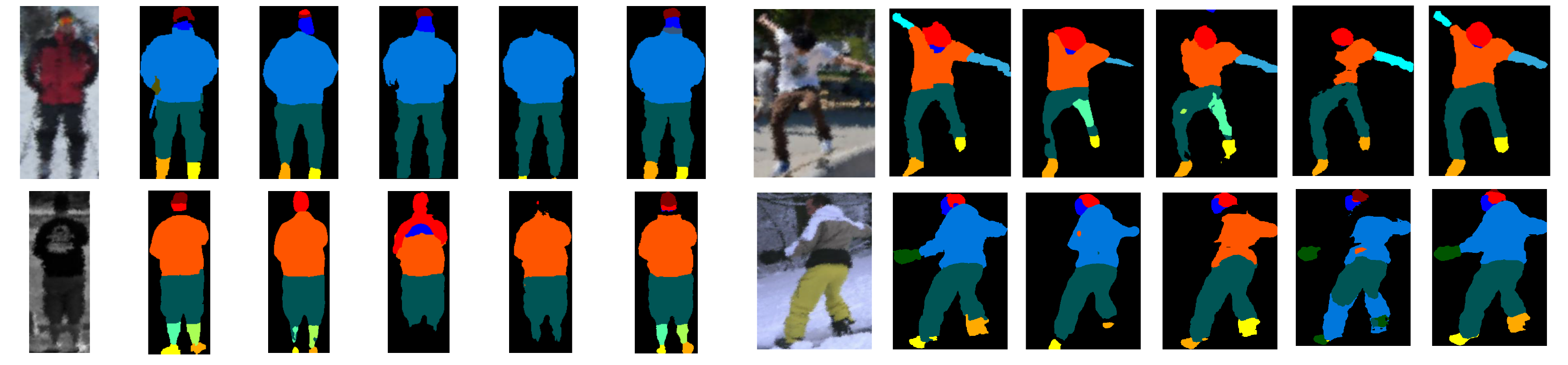} \\
			(c) Glass blur corruption with severity level 3. \\
			\includegraphics[width = 0.85\textwidth]{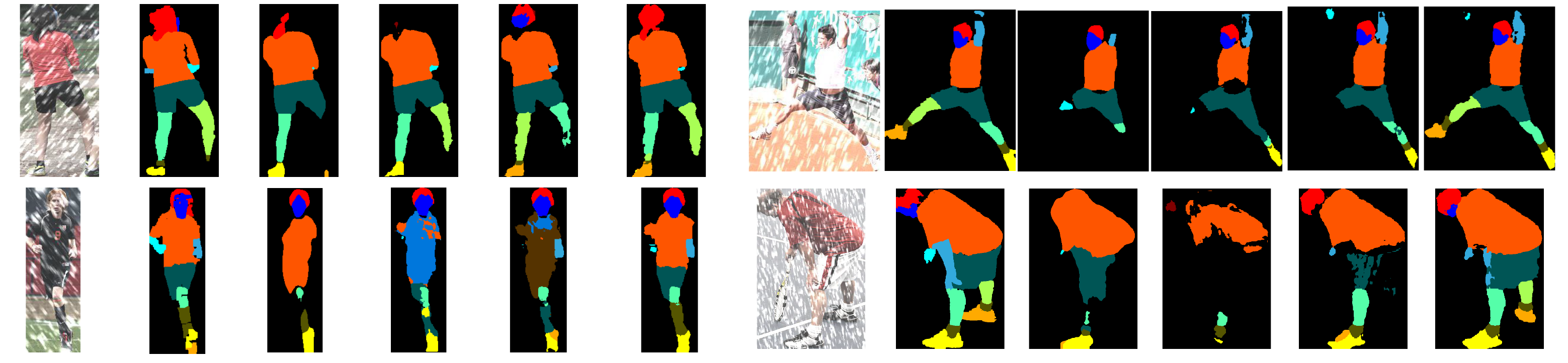} \\
			(d) Snow corruption with severity level 4.  \\
			\includegraphics[width = 0.85\textwidth]{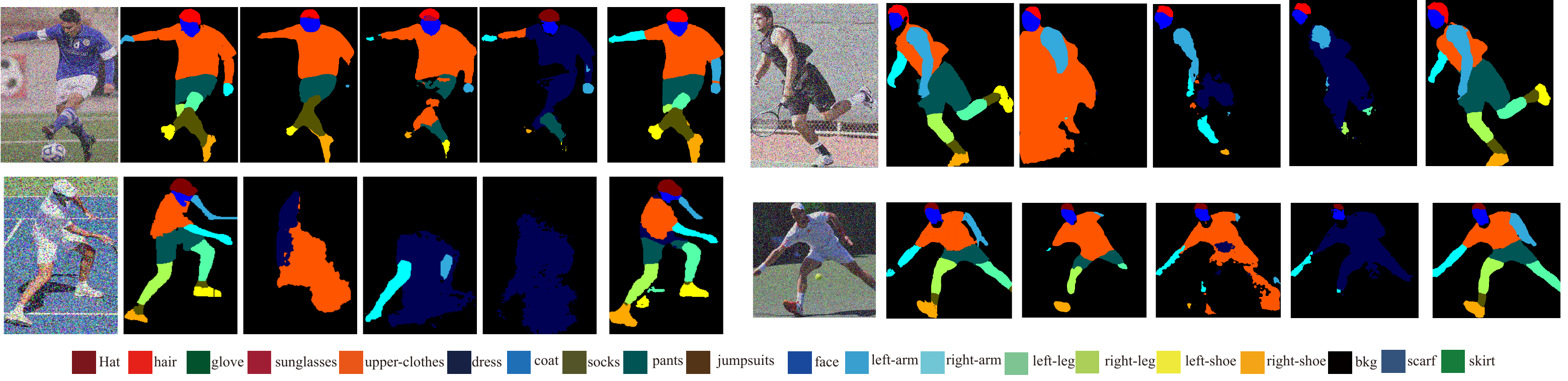} \\
			(e) Impulse noise corruption with severity level 5. \\
		\end{tabular}
	\end{center}
	\caption{Qualitative results on the LIP-C dataset. From left to right are input image, ground-truth, JPPNet \cite{liang2018look}, CE2P \cite{ruan2019devil}, SCHP \cite{li2020self}, and SCHP-HeterAug.
	}
	\label{fig:final}
\end{figure*}

\textbf{Qualitative comparison with the state-of-the-arts.}
To further verify the effectiveness of the proposed heterogeneous augmentation mechanism, we also provide visualization results compared with other state-of-the-art human parsing models. In Figure \ref{fig:final}, we compare the proposed SCHP-HeterAug with three human parsing models, \textit{i.e.}, JPPNet \cite{liang2018look}, CE2P \cite{ruan2019devil} and SCHP \cite{li2020self}, and the results are tested on LIP-C dataset. In Figure \ref{fig:final} (a), if the severity level is small, JPPNet, CE2P, and SCHP  methods predict relatively correct and complete parsing results. As depicted in Figure \ref{fig:final} (d) and (e), as the severity level increases to 4 and 5, these comparison methods cannot parse the input image well. Consequently, they produce imperfect maps, misjudging some regions with background class.
The incomplete parsing maps show that the big severity level hurts the human parsing model more. On the contrary, the proposed SCHP-HeterAug method still achieves relatively accurate and intact parsing maps. Thus, the visualization results
demonstrate that the proposed heterogeneous augmentation mechanism can help the model to improve its robustness while facing various common image corruptions.

\begin{figure}[!th]
  \centering
  \includegraphics[width=0.9\linewidth]{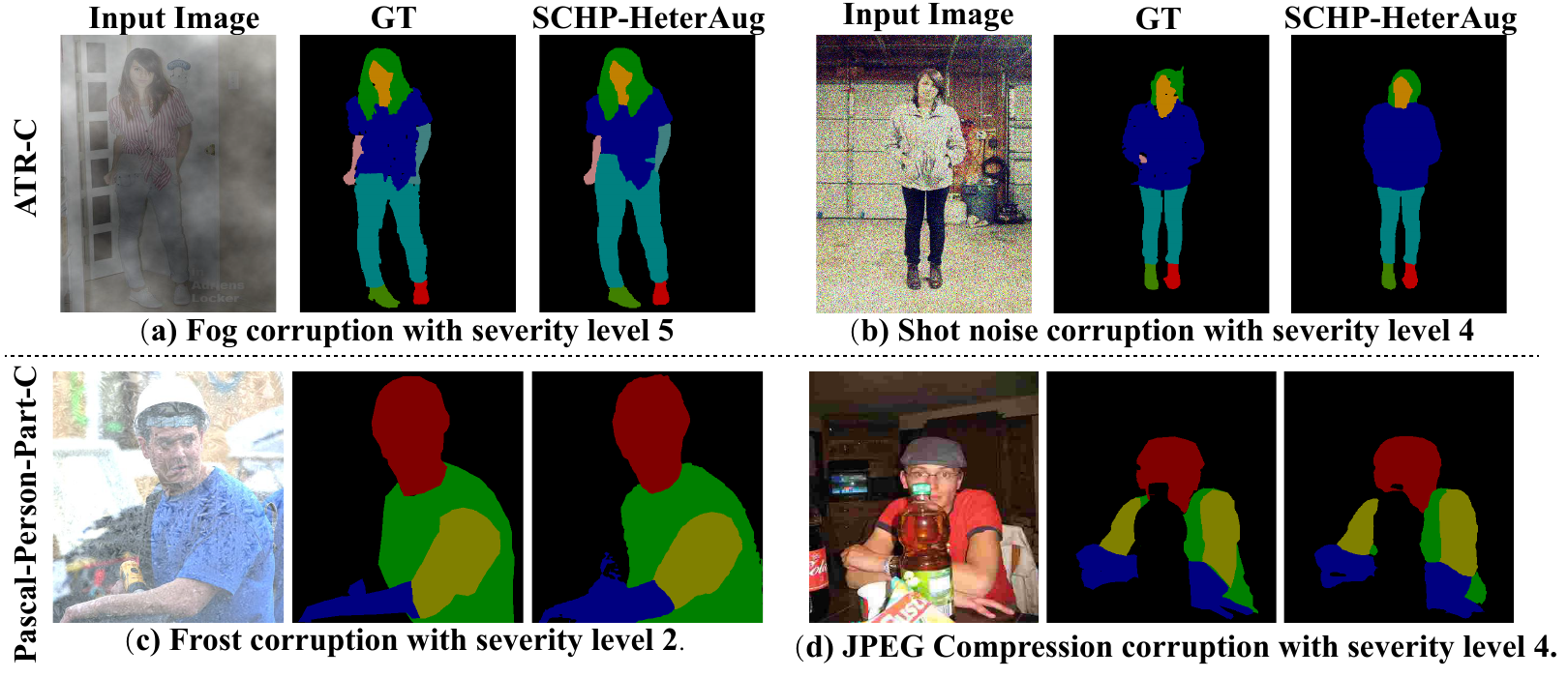}
  \caption{Visualization results on the ATR-C and Pascal-Person-Part-C datasets.
  }
  \label{fig:finalatrpascal}
\end{figure}

\textbf{Visualization results on ATR-C and Pascal-Person-Part-C datasets.}
As shown in Figure \ref{fig:finalatrpascal}, the proposed heterogeneous augmentation mechanism can help the human parser obtain good prediction results while the input images are corrupted with different common corruption operations.

\begin{figure}[h]
  \centering
  \includegraphics[width=0.9\linewidth]{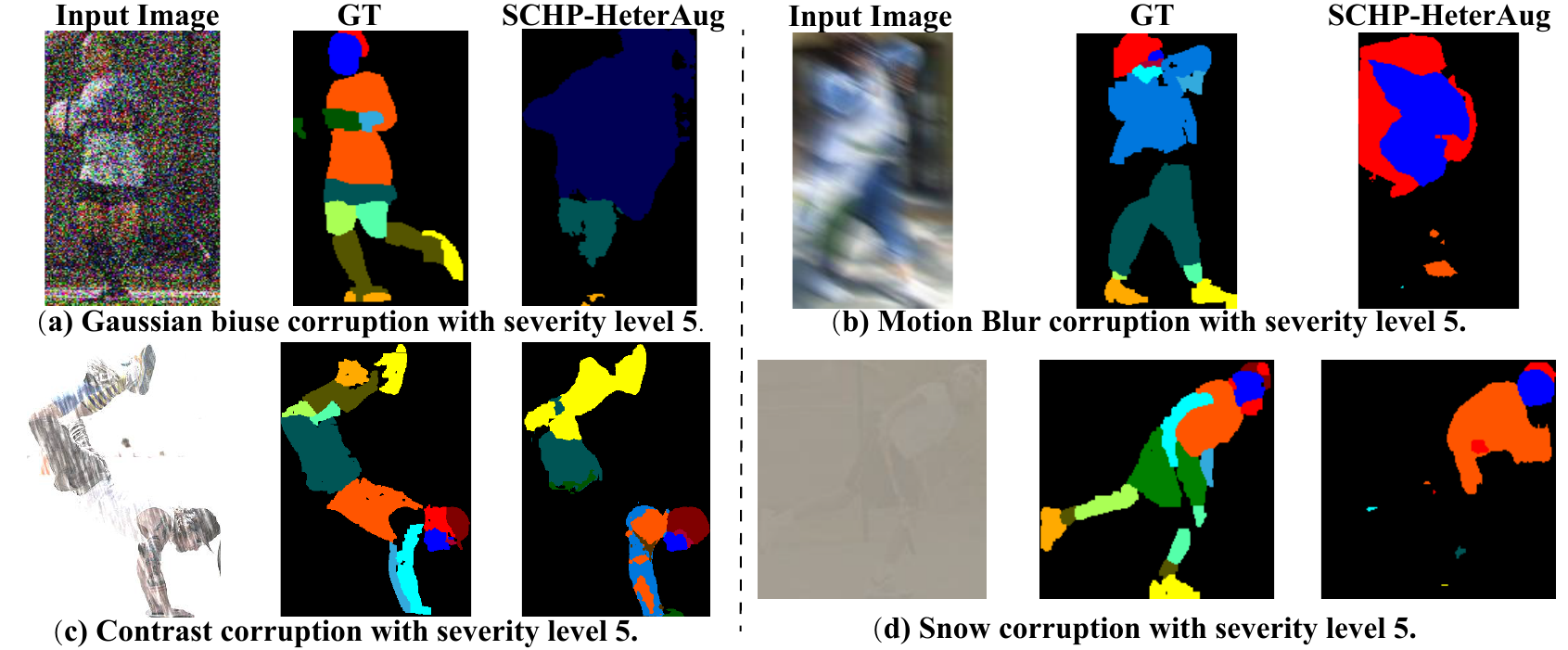}
  \caption{Failure cases on corrupted LIP-C dataset with SCHP-HeterAug.
  }
  \label{fig:error}
\end{figure}

\textbf{Failure cases.} As shown in Figure \ref{fig:error}, while the input images are corrupted with a large severity level, the proposed SCHP-HeterAug method obtains unsatisfactory parsing results. The qualities of all input images are too bad, for example, some regions are similar to the background (Figure \ref{fig:error} (a) and (c)). Additionally, the human regions are extremely blurry (Figure \ref{fig:error} (b)), the human region is hard to be directly identified (Figure \ref{fig:error} (d)). In the future, the robust human parsing solution should take care of these phenomena.

\section{Conclusion}\label{conlusion}
In this paper, we propose a novel robust human parsing benchmark that explores the robustness of human parsers towards common image corruptions. First, we evaluate current state-of-the-art human parsing models on three corrupted human parsing benchmarks, \textit{i.e.}, LIP-C, ATR-C, and Pascal-Person-Part-C. The plummeted performance of the current well-trained human parsers explains the robustness that needs to be further taken into consideration.
To improve the robustness of current human parsers, we propose a novel heterogeneous augmentation mechanism, which combines image-aware and model-aware augmentations in a sequential manner.
Our proposed method seamlessly integrates into any human parsing model with minimal additional computation, yet it delivers substantial performance gains.
In the future, except for the common corruption, we will further explore the influence of adversarial samples on the human parsing task. In addition, current robustness is only explored on the single human parsing datasets, but the multi-person human parsing problem not only contains the pixel-wise classification but also the human bounding box regression. Thus, a suitable mechanism for improving the multi-person human parsing models is also an interesting challenge.

\ifCLASSOPTIONcaptionsoff
  \newpage
\fi

\small
\bibliographystyle{unsrt}
\bibliography{egbib}

\begin{thebibliography}{10}

\bibitem{wang2018attentive}
Wenguan Wang, Yuanlu Xu, Jianbing Shen, and Song-Chun Zhu.
\newblock Attentive fashion grammar network for fashion landmark detection and
  clothing category classification.
\newblock In {\em Proc. of IEEE Conf. CVPR}, pages 4271--4280, 2018.

\bibitem{dong2019towards}
Haoye Dong, Xiaodan Liang, Xiaohui Shen, Bochao Wang, Hanjiang Lai, Jia Zhu,
  Zhiting Hu, and Jian Yin.
\newblock Towards multi-pose guided virtual try-on network.
\newblock In {\em Proc. of IEEE Conf. ICCV}, pages 9026--9035, 2019.

\bibitem{yang2020towards}
Han Yang, Ruimao Zhang, Xiaobao Guo, Wei Liu, Wangmeng Zuo, and Ping Luo.
\newblock Towards photo-realistic virtual try-on by adaptively
  generating-preserving image content.
\newblock In {\em Proc. of IEEE Conf. CVPR}, pages 7850--7859, 2020.

\bibitem{ge2021disentangled}
Chongjian Ge, Yibing Song, Yuying Ge, Han Yang, Wei Liu, and Ping Luo.
\newblock Disentangled cycle consistency for highly-realistic virtual try-on.
\newblock In {\em Proc. of IEEE Conf. CVPR}, pages 16928--16937, 2021.

\bibitem{song2020unpaired}
Sijie Song, Wei Zhang, Jiaying Liu, Zongming Guo, and Tao Mei.
\newblock Unpaired person image generation with semantic parsing
  transformation.
\newblock {\em IEEE Transactions on Pattern Analysis and Machine Intelligence
  (TPAMI)}, 43(11):4161--4176, 2020.

\bibitem{men2020controllable}
Yifang Men, Yiming Mao, Yuning Jiang, Wei-Ying Ma, and Zhouhui Lian.
\newblock Controllable person image synthesis with attribute-decomposed gan.
\newblock In {\em Proc. of IEEE Conf. CVPR}, pages 5084--5093, 2020.

\bibitem{zhu2020identity}
Kuan Zhu, Haiyun Guo, Zhiwei Liu, Ming Tang, and Jinqiao Wang.
\newblock Identity-guided human semantic parsing for person re-identification.
\newblock In {\em Proc. of Conf. ECCV}, pages 346--363. Springer, 2020.

\bibitem{icassp22sgg}
Xuezhi Tong, Rui Wang, Chuan Wang, Sanyi Zhang, and Xiaochun Cao.
\newblock {PMP-NET:} rethinking visual context for scene graph generation.
\newblock In {\em Proc. of Conf. ICASSP}, pages 1940--1944, Virtual and
  Singapore, 2022.

\bibitem{acmmm23psg}
Jiarui Yang, Chuan Wang, Zeming Liu, Jiahong Wu, Dongsheng Wang, Liang Yang,
  and Xiaochun Cao.
\newblock {Focusing on Flexible Masks: } a novel framework for panoptic scene
  graph generation with relation constraints.
\newblock In {\em Proc. of Conf. ACM MM}, pages 1--9, Ottawa, Canada, 2023.

\bibitem{qi2018learning}
Siyuan Qi, Wenguan Wang, Baoxiong Jia, Jianbing Shen, and Song-Chun Zhu.
\newblock Learning human-object interactions by graph parsing neural networks.
\newblock In {\em Proc. of Conf. ECCV}, pages 401--417, 2018.

\bibitem{zhou2021cascaded}
Tianfei Zhou, Siyuan Qi, Wenguan Wang, Jianbing Shen, and Song-Chun Zhu.
\newblock Cascaded parsing of human-object interaction recognition.
\newblock {\em IEEE Transactions on Pattern Analysis and Machine Intelligence
  (TPAMI)}, 2021.

\bibitem{fang2019graspnet}
Hao-Shu Fang, Chenxi Wang, Minghao Gou, and Cewu Lu.
\newblock Graspnet: A large-scale clustered and densely annotated dataset for
  object grasping.
\newblock {\em arXiv preprint arXiv:1912.13470}, 2019.

\bibitem{liu2021super}
Yunan Liu, Shanshan Zhang, Jie Xu, Jian Yang, and Yu-Wing Tai.
\newblock An accurate and lightweight method for human body image
  super-resolution.
\newblock {\em IEEE Transactions on Image Processing (TIP)}, 30:2888--2897,
  2021.

\bibitem{li2020self}
Peike Li, Yunqiu Xu, Yunchao Wei, and Yi~Yang.
\newblock Self-correction for human parsing.
\newblock {\em IEEE Transactions on Pattern Analysis and Machine Intelligence
  (TPAMI)}, 44(6):3260--3271, 2022.

\bibitem{wang2021hierarchical}
Wenguan Wang, Tianfei Zhou, Siyuan Qi, Jianbing Shen, and Song-Chun Zhu.
\newblock Hierarchical human semantic parsing with comprehensive part-relation
  modeling.
\newblock {\em IEEE TPAMI}, 2021.

\bibitem{ruan2019devil}
Tao Ruan, Ting Liu, Zilong Huang, Yunchao Wei, Shikui Wei, and Yao Zhao.
\newblock Devil in the details: Towards accurate single and multiple human
  parsing.
\newblock In {\em Proc. of Conf. AAAI}, volume~33, pages 4814--4821, 2019.

\bibitem{zhang2021human}
Sanyi Zhang, Guo-Jun Qi, Xiaochun Cao, Zhanjie Song, and Jie Zhou.
\newblock Human parsing with pyramidical gather-excite context.
\newblock {\em IEEE Transactions on Circuits and Systems for Video Technology
  (TCSVT)}, 31(3):1016--1030, 2021.

\bibitem{hendrycks2019augmix}
Dan Hendrycks, Norman Mu, Ekin~D Cubuk, Barret Zoph, Justin Gilmer, and Balaji
  Lakshminarayanan.
\newblock Augmix: A simple data processing method to improve robustness and
  uncertainty.
\newblock In {\em Proc. of Conf. ICLR}, 2020.

\bibitem{michaelis2019dragon}
Claudio Michaelis, Benjamin Mitzkus, Robert Geirhos, Evgenia Rusak, Oliver
  Bringmann, Alexander~S. Ecker, Matthias Bethge, and Wieland Brendel.
\newblock Benchmarking robustness in object detection: Autonomous driving when
  winter is coming.
\newblock {\em arXiv preprint arXiv:1907.07484}, 2019.

\bibitem{kamann2020benchmarking}
Christoph Kamann and Carsten Rother.
\newblock Benchmarking the robustness of semantic segmentation models.
\newblock In {\em Proc. of IEEE Conf. CVPR}, pages 8828--8838, 2020.

\bibitem{kamann2020increasing}
Christoph Kamann and Carsten Rother.
\newblock Increasing the robustness of semantic segmentation models with
  painting-by-numbers.
\newblock In {\em Proc. of Conf. ECCV}, pages 369--387, 2020.

\bibitem{wang2021human}
Jiahang Wang, Sheng Jin, Wentao Liu, Weizhong Liu, Chen Qian, and Ping Luo.
\newblock When human pose estimation meets robustness: Adversarial algorithms
  and benchmarks.
\newblock In {\em Proc. of IEEE Conf. CVPR}, pages 11855--11864, 2021.

\bibitem{liang2018look}
Xiaodan Liang, Ke~Gong, Xiaohui Shen, and Liang Lin.
\newblock Look into person: Joint body parsing \& pose estimation network and a
  new benchmark.
\newblock {\em IEEE Transactions on Pattern Analysis and Machine Intelligence
  (TPAMI)}, 41(4):871--885, 2018.

\bibitem{liang2015deep}
Xiaodan Liang, Si~Liu, Xiaohui Shen, Jianchao Yang, Luoqi Liu, Jian Dong, Liang
  Lin, and Shuicheng Yan.
\newblock Deep human parsing with active template regression.
\newblock {\em IEEE Transactions on Pattern Analysis and Machine Intelligence
  (TPAMI)}, 37(12):2402--2414, 2015.

\bibitem{chen2014detect}
Xianjie Chen, Roozbeh Mottaghi, Xiaobai Liu, Sanja Fidler, Raquel Urtasun, and
  Alan Yuille.
\newblock Detect what you can: Detecting and representing objects using
  holistic models and body parts.
\newblock In {\em Proc. of IEEE Conf. CVPR}, pages 1971--1978, 2014.

\bibitem{everingham2010pascal}
Mark Everingham, Luc Van~Gool, Christopher~KI Williams, John Winn, and Andrew
  Zisserman.
\newblock The pascal visual object classes (voc) challenge.
\newblock {\em International Journal of Computer Vision (IJCV)},
  88(2):303--338, 2010.

\bibitem{yamaguchi2012parsing}
Kota Yamaguchi, M~Hadi Kiapour, Luis~E Ortiz, and Tamara~L Berg.
\newblock Parsing clothing in fashion photographs.
\newblock In {\em Proc. of IEEE Conf. CVPR}, pages 3570--3577. IEEE, 2012.

\bibitem{liu2013fashion}
Si~Liu, Jiashi Feng, Csaba Domokos, Hui Xu, Junshi Huang, Zhenzhen Hu, and
  Shuicheng Yan.
\newblock Fashion parsing with weak color-category labels.
\newblock {\em IEEE Transactions on Multimedia (TMM)}, 16(1):253--265, 2014.

\bibitem{bo2011shape}
Yihang Bo and Charless~C Fowlkes.
\newblock Shape-based pedestrian parsing.
\newblock In {\em Proc. of IEEE Conf. CVPR}, pages 2265--2272, 2011.

\bibitem{dong2013deformable}
Jian Dong, Qiang Chen, Wei Xia, Zhongyang Huang, and Shuicheng Yan.
\newblock A deformable mixture parsing model with parselets.
\newblock In {\em Proc. of IEEE Conf. ICCV}, pages 3408--3415, 2013.

\bibitem{yamaguchi2013paper}
Kota Yamaguchi, M~Hadi~Kiapour, and Tamara~L Berg.
\newblock Paper doll parsing: Retrieving similar styles to parse clothing
  items.
\newblock In {\em Proc. of IEEE Conf. ICCV}, pages 3519--3526, 2013.

\bibitem{zhu2008max}
Long Zhu, Yuanhao Chen, Yifei Lu, Chenxi Lin, and Alan Yuille.
\newblock Max margin and/or graph learning for parsing the human body.
\newblock In {\em Proc. of IEEE Conf. CVPR}, pages 1--8, 2008.

\bibitem{liang2015human}
Xiaodan Liang, Chunyan Xu, Xiaohui Shen, Jianchao Yang, Si~Liu, Jinhui Tang,
  Liang Lin, and Shuicheng Yan.
\newblock Human parsing with contextualized convolutional neural network.
\newblock In {\em Proc. of IEEE Conf. ICCV}, pages 1386--1394, 2015.

\bibitem{liang2016semantic}
Xiaodan Liang, Xiaohui Shen, Donglai Xiang, Jiashi Feng, Liang Lin, and
  Shuicheng Yan.
\newblock Semantic object parsing with local-global long short-term memory.
\newblock In {\em Proc. of IEEE Conf. CVPR}, pages 3185--3193, 2016.

\bibitem{liang2016eccv}
Xiaodan Liang, Xiaohui Shen, Jiashi Feng, Liang Lin, and Shuicheng Yan.
\newblock Semantic object parsing with graph lstm.
\newblock In {\em Proc. of Conf. ECCV}, pages 125--143, 2016.

\bibitem{luo2013pedestrian}
Ping Luo, Xiaogang Wang, and Xiaoou Tang.
\newblock Pedestrian parsing via deep decompositional network.
\newblock In {\em Proc. of IEEE Conf. ICCV}, pages 2648--2655, 2013.

\bibitem{wang2021exploring}
Wenguan Wang, Tianfei Zhou, Fisher Yu, Jifeng Dai, Ender Konukoglu, and Luc
  Van~Gool.
\newblock Exploring cross-image pixel contrast for semantic segmentation.
\newblock In {\em Proc. of IEEE Conf. ICCV}, pages 7303--7313, 2021.

\bibitem{zhou2022rethinking}
Tianfei Zhou, Wenguan Wang, Ender Konukoglu, and Luc Van~Gool.
\newblock Rethinking semantic segmentation: A prototype view.
\newblock In {\em Proc. of IEEE Conf. CVPR}, pages 2582--2593, 2022.

\bibitem{zhang2020CVPR}
Ziwei Zhang, Chi Su, Liang Zheng, and Xiaodong Xie.
\newblock Correlating edge, pose with parsing.
\newblock In {\em Proc. of IEEE Conf. CVPR}, pages 8900--8909, 2020.

\bibitem{zhou2021differentiable}
Tianfei Zhou, Wenguan Wang, Si~Liu, Yi~Yang, and Luc Van~Gool.
\newblock Differentiable multi-granularity human representation learning for
  instance-aware human semantic parsing.
\newblock In {\em Proc. of IEEE Conf. CVPR}, pages 1622--1631, 2021.

\bibitem{wang2020paying}
Wenguan Wang, Jianbing Shen, Xiankai Lu, Steven~CH Hoi, and Haibin Ling.
\newblock Paying attention to video object pattern understanding.
\newblock {\em IEEE Transactions on Pattern Analysis and Machine Intelligence
  (TPAMI)}, 43(7):2413--2428, 2020.

\bibitem{ji2019learning}
Ruyi Ji, Dawei Du, Libo Zhang, Longyin Wen, Yanjun Wu, Chen Zhao, Feiyue Huang,
  and Siwei Lyu.
\newblock Learning semantic neural tree for human parsing.
\newblock In {\em Proc. of Conf. ECCV}, pages 205--221, 2020.

\bibitem{wang2019learning}
Wenguan Wang, Zhijie Zhang, Siyuan Qi, Jianbing Shen, Yanwei Pang, and Ling
  Shao.
\newblock Learning compositional neural information fusion for human parsing.
\newblock In {\em Proc. of IEEE Conf. ICCV}, pages 5703--5713, 2019.

\bibitem{wang2020hierarchical}
Wenguan Wang, Hailong Zhu, Jifeng Dai, Yanwei Pang, Jianbing Shen, and Ling
  Shao.
\newblock Hierarchical human parsing with typed part-relation reasoning.
\newblock In {\em Proc. of IEEE Conf. CVPR}, pages 8929--8939, 2020.

\bibitem{gong2019graphonomy}
Ke~Gong, Yiming Gao, Xiaodan Liang, Xiaohui Shen, Meng Wang, and Liang Lin.
\newblock Graphonomy: Universal human parsing via graph transfer learning.
\newblock In {\em Proc. of IEEE Conf. CVPR}, pages 7450--7459, 2019.

\bibitem{he2019grapy}
Haoyu He, Jing Zhang, Qiming Zhang, and Dacheng Tao.
\newblock Grapy-ml: Graph pyramid mutual learning for cross-dataset human
  parsing.
\newblock In {\em Proc. of Conf. AAAI}, volume~34, pages 10949--10956, 2020.

\bibitem{li2022deep}
Liulei Li, Tianfei Zhou, Wenguan Wang, Jianwu Li, and Yi~Yang.
\newblock Deep hierarchical semantic segmentation.
\newblock In {\em Proc. of IEEE Conf. CVPR}, pages 1246--1257, 2022.

\bibitem{luo2018macro}
Yawei Luo, Zhedong Zheng, Liang Zheng, Tao Guan, Junqing Yu, and Yi~Yang.
\newblock Macro-micro adversarial network for human parsing.
\newblock In {\em Proc. of Conf. ECCV}, pages 418--434, 2018.

\bibitem{liu2018cross}
Si~Liu, Yao Sun, Defa Zhu, Guanghui Ren, Yu~Chen, Jiashi Feng, and Jizhong Han.
\newblock Cross-domain human parsing via adversarial feature and label
  adaptation.
\newblock In {\em Proc. of Conf. AAAI}, volume~32, 2018.

\bibitem{zhou2022consistency}
Tao Zhou, Huazhu Fu, Chen Gong, Ling Shao, Fatih Porikli, Haibin Ling, and
  Jianbing Shen.
\newblock Consistency and diversity induced human motion segmentation.
\newblock {\em IEEE Transactions on Pattern Analysis and Machine Intelligence
  (TPAMI)}, 45(1):197--210, 2022.

\bibitem{li2020selfsemi}
Tao Li, Zhiyuan Liang, Sanyuan Zhao, Jiahao Gong, and Jianbing Shen.
\newblock Self-learning with rectification strategy for human parsing.
\newblock In {\em Proc. of IEEE Conf. CVPR}, pages 9263--9272, 2020.

\bibitem{wang2021augmax}
Haotao Wang, Chaowei Xiao, Jean Kossaifi, Zhiding Yu, Anima Anandkumar, and
  Zhangyang Wang.
\newblock Augmax: Adversarial composition of random augmentations for robust
  training.
\newblock In {\em Proc. of Conf. NeurIPS}, volume~34, 2021.

\bibitem{jiang2020robust}
Ziyu Jiang, Tianlong Chen, Ting Chen, and Zhangyang Wang.
\newblock Robust pre-training by adversarial contrastive learning.
\newblock In {\em Proc. of Conf. NeurIPS}, volume~33, pages 16199--16210, 2020.

\bibitem{weng2018evaluating}
Tsui-Wei Weng, Huan Zhang, Pin-Yu Chen, Jinfeng Yi, Dong Su, Yupeng Gao,
  Cho-Jui Hsieh, and Luca Daniel.
\newblock Evaluating the robustness of neural networks: An extreme value theory
  approach.
\newblock In {\em Proc. of Conf. ICLR}, 2018.

\bibitem{zheng2016improving}
Stephan Zheng, Yang Song, Thomas Leung, and Ian Goodfellow.
\newblock Improving the robustness of deep neural networks via stability
  training.
\newblock In {\em Proc. of IEEE Conf. CVPR}, pages 4480--4488, 2016.

\bibitem{hendrycks2021many}
Dan Hendrycks, Steven Basart, Norman Mu, Saurav Kadavath, Frank Wang, Evan
  Dorundo, Rahul Desai, Tyler Zhu, Samyak Parajuli, Mike Guo, et~al.
\newblock The many faces of robustness: A critical analysis of
  out-of-distribution generalization.
\newblock In {\em Proc. of IEEE Conf. ICCV}, pages 8340--8349, 2021.

\bibitem{kamann2021benchmarking}
Christoph Kamann and Carsten Rother.
\newblock Benchmarking the robustness of semantic segmentation models with
  respect to common corruptions.
\newblock {\em International Journal of Computer Vision (IJCV)},
  129(2):462--483, 2021.

\bibitem{xu2021dynamic}
Xiaogang Xu, Hengshuang Zhao, and Jiaya Jia.
\newblock Dynamic divide-and-conquer adversarial training for robust semantic
  segmentation.
\newblock In {\em Proc. of IEEE Conf. ICCV}, pages 7486--7495, 2021.

\bibitem{cubuk2018autoaugment}
Ekin~D Cubuk, Barret Zoph, Dandelion Mane, Vijay Vasudevan, and Quoc~V Le.
\newblock Autoaugment: Learning augmentation policies from data.
\newblock In {\em Proc. of IEEE Conf. CVPR}, pages 113--123, 2019.

\bibitem{gao2019res2net}
Shanghua Gao, Ming-Ming Cheng, Kai Zhao, Xin-Yu Zhang, Ming-Hsuan Yang, and
  Philip~HS Torr.
\newblock Res2net: A new multi-scale backbone architecture.
\newblock {\em IEEE Transactions on Pattern Analysis and Machine Intelligence
  (TPAMI)}, 43(2):652--662, 2019.

\bibitem{WangSCJDZLMTWLX19}
Jingdong Wang, Ke~Sun, Tianheng Cheng, Borui Jiang, Chaorui Deng, Yang Zhao,
  Dong Liu, Yadong Mu, Mingkui Tan, Xinggang Wang, Wenyu Liu, and Bin Xiao.
\newblock Deep high-resolution representation learning for visual recognition.
\newblock {\em IEEE Transactions on Pattern Analysis and Machine Intelligence
  (TPAMI)}, 43(10):3349--3364, 2020.

\bibitem{YuanCW19}
Yuhui Yuan, Xilin Chen, and Jingdong Wang.
\newblock Object-contextual representations for semantic segmentation.
\newblock In {\em Proc. of Conf. ECCV}, pages 173--190, 2020.

\bibitem{Luo2018TGPnet}
Xianghui Luo, Zhuo Su, Jiaming Guo, Gengwei Zhang, and Xiangjian He.
\newblock Trusted guidance pyramid network for human parsing.
\newblock In {\em Proc. of Conf. ACM MM}, pages 654--662, 2018.

\bibitem{chen2020gridmask}
Pengguang Chen, Shu Liu, Hengshuang Zhao, and Jiaya Jia.
\newblock Gridmask data augmentation.
\newblock {\em arXiv preprint arXiv:2001.04086}, 2020.

\bibitem{zhao2017pyramid}
Hengshuang Zhao, Jianping Shi, Xiaojuan Qi, Xiaogang Wang, and Jiaya Jia.
\newblock Pyramid scene parsing network.
\newblock In {\em Proc. IEEE Conf. CVPR}, pages 2881--2890, 2017.

\bibitem{madry2017towards}
Aleksander Madry, Aleksandar Makelov, Ludwig Schmidt, Dimitris Tsipras, and
  Adrian Vladu.
\newblock Towards deep learning models resistant to adversarial attacks.
\newblock In {\em Proc. of Conf. ICLR}, 2018.

\bibitem{chen2018encoder}
Liang-Chieh Chen, Yukun Zhu, George Papandreou, Florian Schroff, and Hartwig
  Adam.
\newblock Encoder-decoder with atrous separable convolution for semantic image
  segmentation.
\newblock In {\em Proc. of Conf. ECCV}, pages 801--818, 2018.

\end{thebibliography}

%

\end{document}